\documentclass{article}

\usepackage{microtype}
\usepackage{graphicx}
\usepackage{subcaption}
\usepackage{booktabs} 

\usepackage{hyperref}
\usepackage{adjustbox}
\usepackage{caption}

\usepackage[most]{tcolorbox}

\usepackage[preprint]{icml2026}

\usepackage{amsmath}
\usepackage{amssymb}
\usepackage{mathtools}
\usepackage{amsthm}
\usepackage{booktabs}
\usepackage{multirow}
\usepackage[table]{xcolor}
\usepackage{makecell}
\usepackage{algorithm}
\usepackage{algorithmic}
\usepackage{amsmath} 

\definecolor{myred}{HTML}{D73027}  
\definecolor{mygreen}{HTML}{1A9850} 

\usepackage[capitalize,noabbrev]{cleveref}

\theoremstyle{plain}

\theoremstyle{definition}

\theoremstyle{remark}

\usepackage[textsize=tiny]{todonotes}

\usepackage{color}
\definecolor{green}{rgb}{0, 0.5, 0}
\definecolor{orange}{rgb}{0.8, 0.6, 0.2}
\definecolor{orange2}{rgb}{1.0, 0.6, 0.2}
\definecolor{red}{rgb}{1.0, 0.0, 0.0}
\definecolor{teal}{rgb}{0.0, 0.4, 0.4}
\definecolor{purple}{rgb}{0.65,0,0.65}
\definecolor{saffron}{rgb}{0.95,0.75,0.2}
\definecolor{turquoise}{rgb}{0.0,0.5,0.5}
\definecolor{black}{rgb}{0.0, 0.0, 0.0}
\definecolor{gray}{rgb}{0.5, 0.5, 0.5}
\hypersetup{
  colorlinks   = true,    
  urlcolor     = purple,    
  linkcolor    = purple,    
  citecolor    = purple      
}

\icmltitlerunning{Submission and Formatting Instructions for ICML 2026}

\begin{document}

\twocolumn[
  \icmltitle{Allocentric Perceiver: Disentangling Allocentric Reasoning from Egocentric Visual Priors via Frame Instantiation}



  \icmlsetsymbol{equal}{*}

  \begin{icmlauthorlist}
    \icmlauthor{Hengyi Wang}{equal,USTC}
    \icmlauthor{Ruiqiang Zhang}{equal,USTC}
    \icmlauthor{Chang Liu}{USTC}
    \icmlauthor{Guanjie Wang}{USTC}
    \icmlauthor{Zehua Ma}{USTC}
    \icmlauthor{Han Fang}{NUS}
    \icmlauthor{Weiming Zhang}{USTC}
  \end{icmlauthorlist}

  \icmlaffiliation{USTC}{Anhui Province Key Laboratory of Digital Security, University of Science and Technology of China}
  \icmlaffiliation{NUS}{National University of Singapore}

  \icmlcorrespondingauthor{Zehua Ma}{mzh045@ustc.edu.cn}
  \icmlcorrespondingauthor{Weiming Zhang}{zhangwm@ustc.edu.cn}

  \icmlkeywords{Machine Learning, ICML}

  \vskip 0.3in
]



\printAffiliationsAndNotice{}  

\begin{abstract}
With the rising need for spatially grounded tasks such as Vision-Language Navigation/Action, allocentric perception capabilities in Vision-Language Models (VLMs) are receiving growing focus. However, VLMs remain brittle on allocentric spatial queries that require explicit perspective shifts, where the answer depends on reasoning in a target-centric frame rather than the observed camera view. Thus, we introduce \textbf{Allocentric Perceiver}, a training-free strategy that recovers metric 3D states from one or more images with off-the-shelf geometric experts, and then instantiates a query-conditioned allocentric reference frame aligned with the instruction’s semantic intent. By deterministically transforming reconstructed geometry into the target frame and prompting the backbone VLM with structured, geometry-grounded representations, Allocentric Perceriver offloads mental rotation from implicit reasoning to explicit computation. We evaluate Allocentric Perciver across multiple backbone families on spatial reasoning benchmarks, observing consistent and substantial gains ($\sim$\textbf{10\%}) on allocentric tasks while maintaining strong egocentric performance, and surpassing both spatial‑perception-finetuned models and state‑of‑the‑art open‑source and proprietary models.  
\end{abstract}

\section{Introduction \label{Intro}}

\begin{figure}[t]
  \centering
  \includegraphics[width=\linewidth]{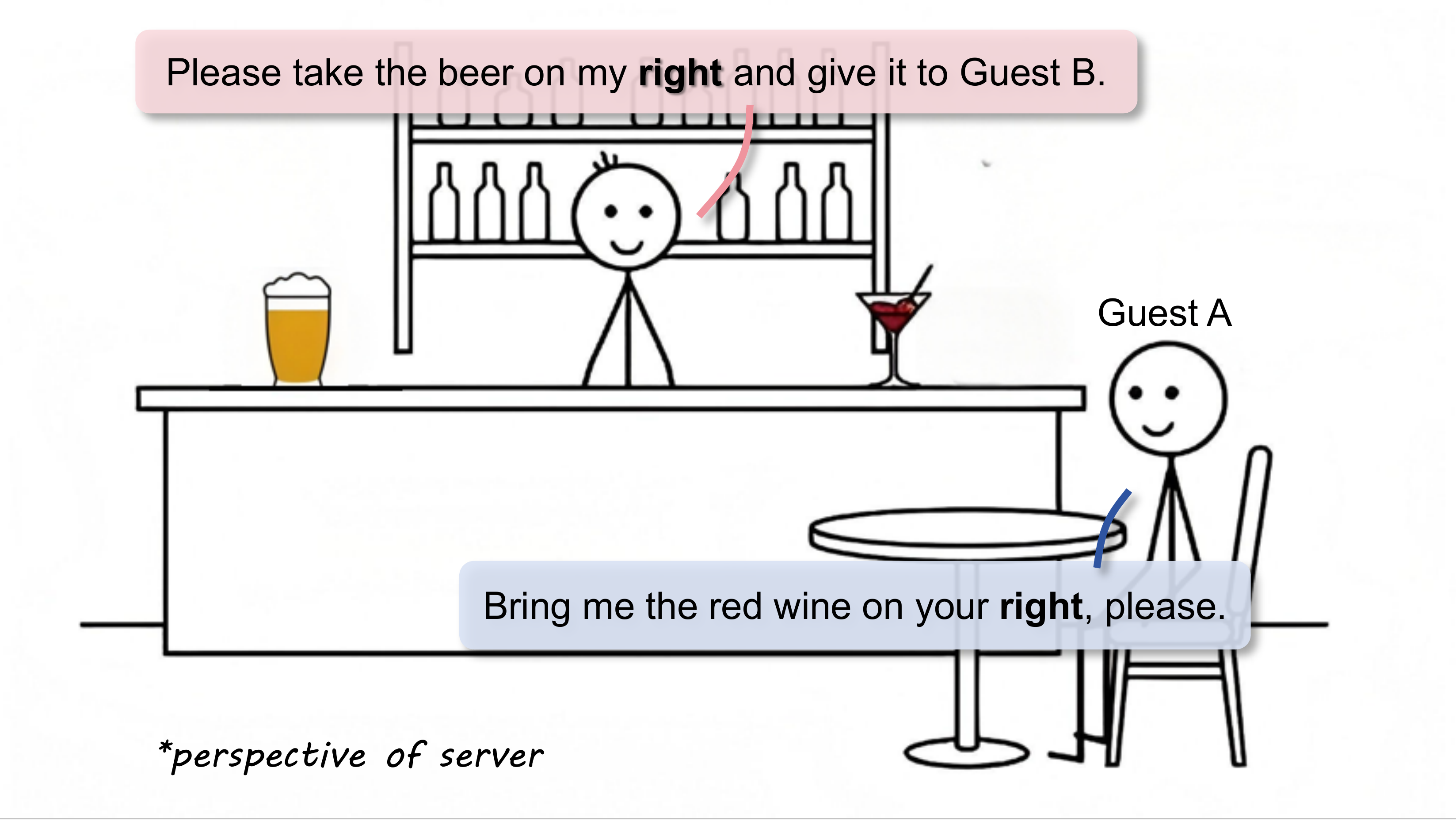}
  \caption{Egocentric V.S. allocentric instructions.}
  \label{server_in_bar}
  \vskip -0.15in
\end{figure}

The pursuit of embodied Artificial Intelligence has long aimed to endow Vision-Language Models (VLMs) with human-like spatial perception, a capability essential for mastering complex open-world tasks, from autonomous driving and outdoor navigation to service robotics in indoor environments. In human biological systems, robust spatial cognition relies on two distinct yet complementary neural coding systems: the \textbf{Egocentric} and the \textbf{Allocentric} reference frames \cite{o1978hippocampus,klatzky1998allocentric}. The \textbf{egocentric} system, primarily associated with the parietal cortex, encodes spatial information relative to the observer, facilitating immediate sensorimotor actions and real-time interaction \cite{goodale1992separate}. Conversely, the \textbf{allocentric} system, rooted in the hippocampal formation, constructs a world-centered ``cognitive map" independent of the observer’s current position. For instance, as in Fig. \ref{server_in_bar}, in a bar, a human waiter can process an immediate egocentric request from guests (``Bring me the red wine on your right") or a complex allocentric instruction from a bartender (``Please take the beer on my right") with equal fluency. 
These two systems together constitute humanity's robust and accurate spatial reasoning capabilities.

While early VLM research primarily focused on image captioning and Visual Question Answering (VQA) tasks \cite{li2022blipbootstrappinglanguageimagepretraining,chen2024rightwayevaluatinglarge,liu2024mmbenchmultimodalmodelallaround}, the surging demand for Vision-Language Navigation \cite{cheng2025navilaleggedrobotvisionlanguageaction,goetting2024endtoendnavigationvisionlanguage} and Vision-Language-Action \cite{brohan2023rt2visionlanguageactionmodelstransfer,kim2024openvlaopensourcevisionlanguageactionmodel} systems has highlighted the need for stronger spatial reasoning capabilities. 

Recent methods have endeavored to equip VLMs with spatial perception and reasoning capabilities, primarily by constructing high-quality VQA datasets grounded in 2D and depth information, integrating specialized spatial-aware modules \cite{bigverdi2024perceptiontokensenhancevisual,liu2025ssrenhancingdepthperception,cheng2024spatialrgptgroundedspatialreasoning,cai2025spatialbotprecisespatialunderstanding,chen2025sdvlmspatialmeasuringunderstanding}, or applying targeted post-training \cite{chen2024spatialvlmendowingvisionlanguagemodels,li2025spatialladderprogressivetrainingspatial,liu2025spatialssrlenhancingspatialunderstanding,batra2025spatialthinkerreinforcing3dreasoning}. Besides, heuristic approaches \cite{li2025seetrektrainingfreespatialprompting,ma2024spatialpinenhancingspatialreasoning,han2025tigertoolintegratedgeometricreasoning,yang2025mindjourneytesttimescalingworld,liu2025abstract3dperceptionspatial} have been proposed to augment these abilities via external expert models.

Even when models are trained or augmented for ``spatial reasoning", the resulting gains are typically most visible on camera-view tasks, and failures persist on problems that require explicit perspective shifts \cite{ma2025spatialreasonerexplicitgeneralizable3d,góral2024seeingeyesevaluatingvisual,zhang2025visionlanguagemodelsrepresentspace}. This deficiency is critical, as allocentric perception is foundational for constructing persistent cognitive maps, enabling embodied navigation, and facilitating robot manipulation \cite{jeffery_mosaic_2024,cartillier2021semanticmapnetbuildingallocentric,chapin2025objectcentricrepresentationsimprovepolicy,zheng_survey_2025}.

\begin{figure}[t]
  \centering
  \includegraphics[width=\linewidth]{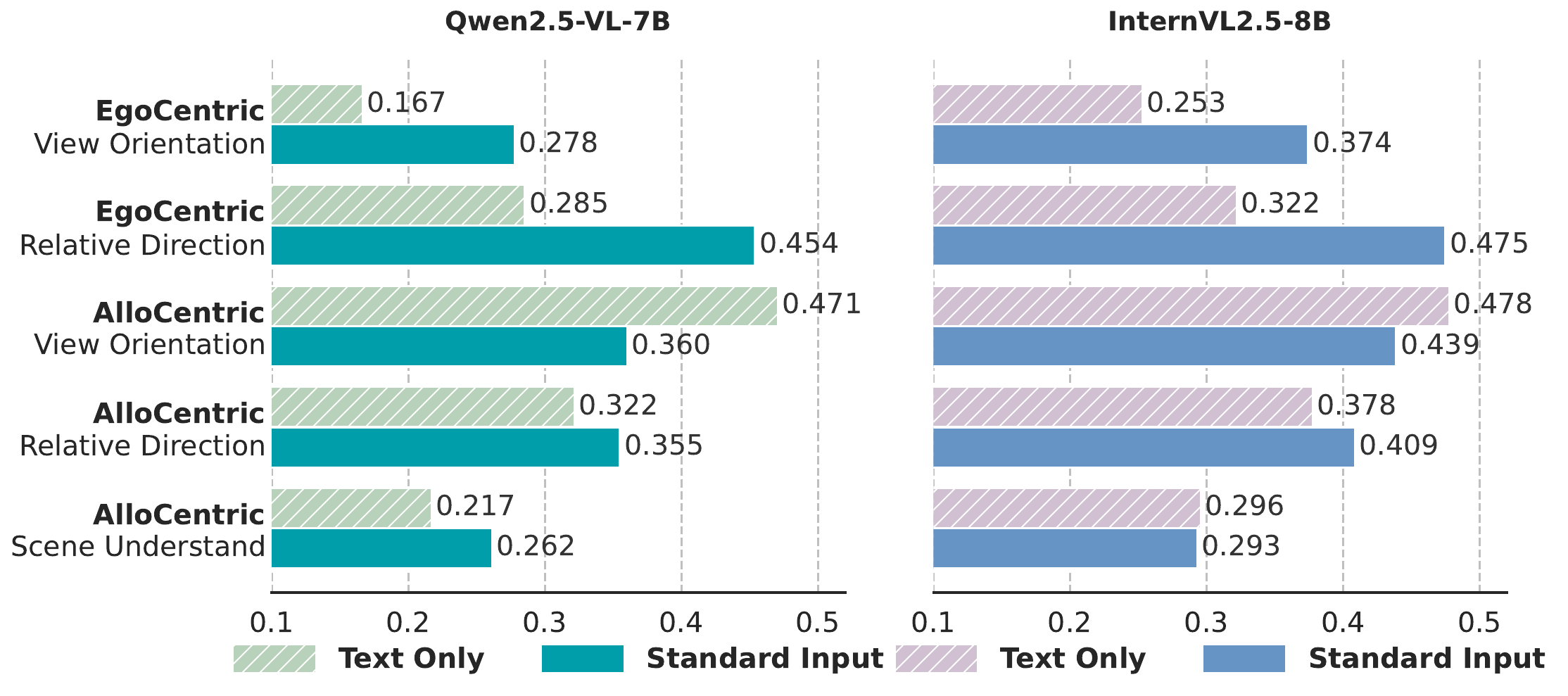}
  \caption{The viewspatial-bench comprises two egocentric tasks and three allocentric tasks. ``Stander input" denotes standard multimodal input, while ``Text only" refers to text-only question inputs. }
  \label{Feasible_study}
  \vskip -0.15in
\end{figure}

To understand the root of this deficiency, we conducted a feasibility study comparing VLM performance on spatial tasks with and without visual input. Specifically, we evaluated the accuracy of Qwen2.5VL-7B \cite{bai2025qwen25vltechnicalreport} and InternVL2.5-8B \cite{chen2025expandingperformanceboundariesopensource} on the ViewSpatial-Bench \cite{li2025viewspatialbench} under two conditions: standard input and text-only input (where images are withheld). The results, shown in Fig. \ref{Feasible_study}, reveal a striking divergence. For Qwen2.5VL-7B, removing visual input led to a sharp \textbf{14.80\%} decline in egocentric task performance; conversely, allocentric performance unexpectedly improved by 1.12\%. InternVL2.5-8B also exhibits a similar phenomenon.

\begin{figure}[t]
  \centering
  \includegraphics[width=0.94\linewidth]{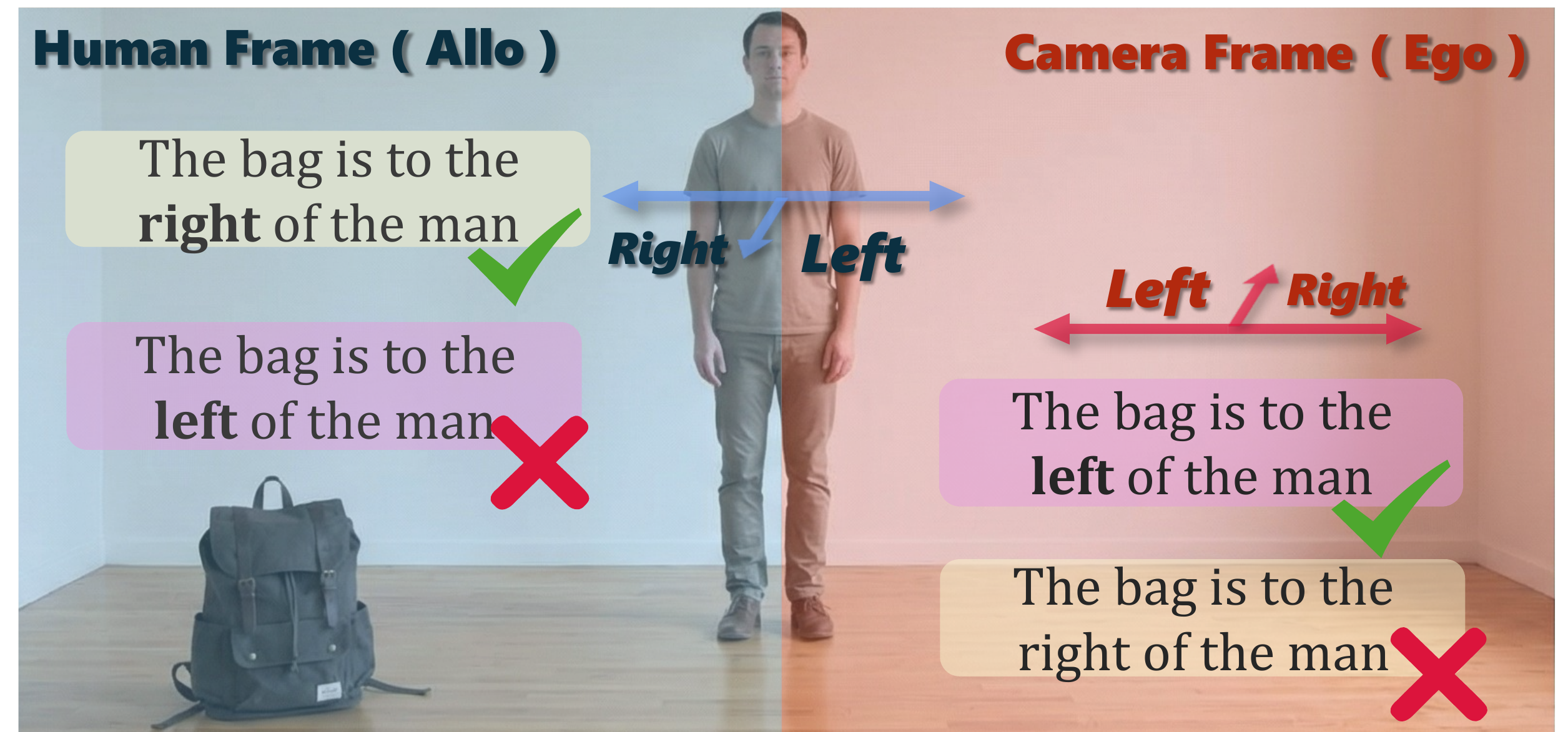}
  \caption{Illustration of Visual-Semantic Ambiguity. The validity of spatial descriptions is contingent upon the reference frame.}
  \label{fig3}
  \vskip -0.15in
\end{figure}

We attribute this phenomenon to the \textbf{inherent perspective ambiguity} in the massive image-text pairs used for training. As detailed in Fig. \ref{fig3}, consider a scene with two conflicting descriptions: \textit{$C_1$: ``The bag is to the left of the man"} and \textit{$C_2$: ``The bag is to the right of the man.''} Under the camera frame (egocentric context), spatial relations are mapped directly to the 2D pixel space; thus, $C_1$ is correct. 
Conversely, under the human frame (allocentric context), spatial relations are defined by the person's orientation; here, $C_2$ is correct. 
Critically, training datasets rarely distinguish these perspectives, introducing a fundamental Visual-Semantic Ambiguity into the model’s learning of spatial relations. This creates the antagonistic relationship observed in our feasibility study. When the model is tasked with allocentric reasoning (e.g., judging position relative to the man), the strong visual prior from the egocentric view (seeing the bag on the left pixel field) generates an adversarial signal that confuses the model's internal ``perspective shift" process.

Motivated by these findings, we propose \textbf{Allocentric Perceiver}, a novel framework designed for allocentric reasoning by mimicking the human cognitive process, including three stages: (1) \textit{Metric-Aware Egocentric Perception}: To acquire target objects' states, we first leverage visual expert models \cite{langsam,ravi2024sam2} head pose estimators \cite{10477888}, and off-the-shelf 3D estimators \cite{lin2025depth3recoveringvisual} to recover the interpretable 3D spatial pose and position of the objects from 2D inputs. (2) \textit{Dynamic Frame Instantiation}: To simulate an explicit perspective shift, we designate the target object as the new coordinate anchor and mathematically formalize the transformation from the egocentric camera view to the target's allocentric reference frame. (3) \textit{Symbolic Geometry Reasoning}: Finally, to reason about surroundings relative in the new frame, we discard raw images and prompt the VLM solely with unambiguous, geometry-grounded textual representations. Our contributions are summarized as follows:

\begin{itemize}
\item We identify a fundamental \textbf{Reference Frame Gap} in VLM spatial reasoning: egocentric visual priors inherently conflict with allocentric queries due to visual-semantic ambiguity in training data, as evidenced by our feasibility study, along with experiments and case analysis over different benchmarks.

\item We propose \textbf{Allocentric Perceiver}, a training-free framework that decouples allocentric reasoning from egocentric visual priors via metric-aware perception and dynamic frame instantiation. This approach enables VLMs to reason allocentric tasks through frame-aligned and geometric grounded prompts.

\item We demonstrate consistent, backbone-agnostic performance gains across multiple multi-perspective benchmarks. Experimental results show that \textbf{Allocentric Perceiver} substantially enhances allocentric reasoning accuracy while bringing a certain degree of improvement to the model's inherent egocentric competence.
\end{itemize}

\section{Related Works}

\subsection{Training VLMs for Spatial Reasoning}

A growing body of work enhances spatial reasoning by training VLMs with targeted data, objectives, or rewards. These approaches often curate instruction-tuning corpora focused on spatial relations \cite{tang2025sparklemasteringbasicspatial,chen2024spatialvlmendowingvisionlanguagemodels}, or inject geometric inductive biases, such as explicit depth encoding, to reduce 2D heuristic reliance \cite{liu2025ssrenhancingdepthperception,chen2025sdvlmspatialmeasuringunderstanding}. More recent efforts explore progressive curricula and reward shaping to foster 3D-consistent reasoning \cite{li2025spatialladderprogressivetrainingspatial,batra2025spatialthinkerreinforcing3dreasoning,ma2025spatialreasonerexplicitgeneralizable3d}.
Despite these gains, such methods frequently exhibit \textit{frame dependence}: performance is robust primarily when queries align with camera or dataset conventions but degrades under perspective shifts or ambiguous references \cite{zhang2025visionlanguagemodelsrepresentspace,li2025viewspatialbench,ma20253dsrbenchcomprehensive3dspatial}. In contrast, we address this reference-frame gap directly by explicitly instantiating a query-conditioned allocentric frame and performing geometric transformations before reasoning.

\subsection{Training-Free Spatial Reasoning via Prompting and Tool Use}

Other training-free methods enhance spatial reasoning by decomposing tasks into intermediate geometric steps, such as coordinate extraction, viewpoint rewriting, or iterative verification. Prompting-centric approaches utilize structured subproblems or perspective-taking prompts, often integrated with self-consistency loops \cite{li2025seetrektrainingfreespatialprompting,ma2024spatialpinenhancingspatialreasoning}. Tool-augmented systems further leverage external modules, such as depth estimators and planners, to provide structured intermediate states for VLM reasoning \cite{han2025tigertoolintegratedgeometricreasoning,yang2025mindjourneytesttimescalingworld,liu2025abstract3dperceptionspatial}. Our approach follows this training-free paradigm but introduces a key distinction: rather than requiring the VLM to perform implicit mental rotation within the language space, we recover metric 3D states in a unified world space and apply query-conditioned transformations to an allocentric frame. This reduces the final reasoning step to a straightforward, geometry-grounded inference.

\subsection{3D-Aware Multimodal Large Language Models}

Recent progress in feed-forward 3D geometry foundation models \cite{wang2025vggtvisualgeometrygrounded,wang2025pi3permutationequivariantvisualgeometry,lin2025depth3recoveringvisual} represented by VGGT, has enabled the recovery of metric 3D structures from unconstrained RGB images, providing geometrically consistent perception for spatial reasoning. 3D-LLM \cite{hong20233dllminjecting3dworld} and PointLLM \cite{xu2024pointllmempoweringlargelanguage} pioneered the direct encoding of 3D point clouds, while more recent generalist frameworks like 3D-LLaVA \cite{deng20253dllavageneralist3dlmms} and 3UR-LLM \cite{xiong20253urllmendtoendmultimodallarge} utilize end-to-end architectures for unified 3D understanding. However, despite access to high-fidelity 3D data, these approaches predominantly reason within a \textit{static global coordinate system}. They lack the mechanisms for dynamic \textit{perspective-taking}, the ability to shift from holistic scene representations to the specific, object-centric reference frames essential for embodied allocentric reasoning.

\begin{figure*}[t]
  \centering
  \includegraphics[width=\textwidth]{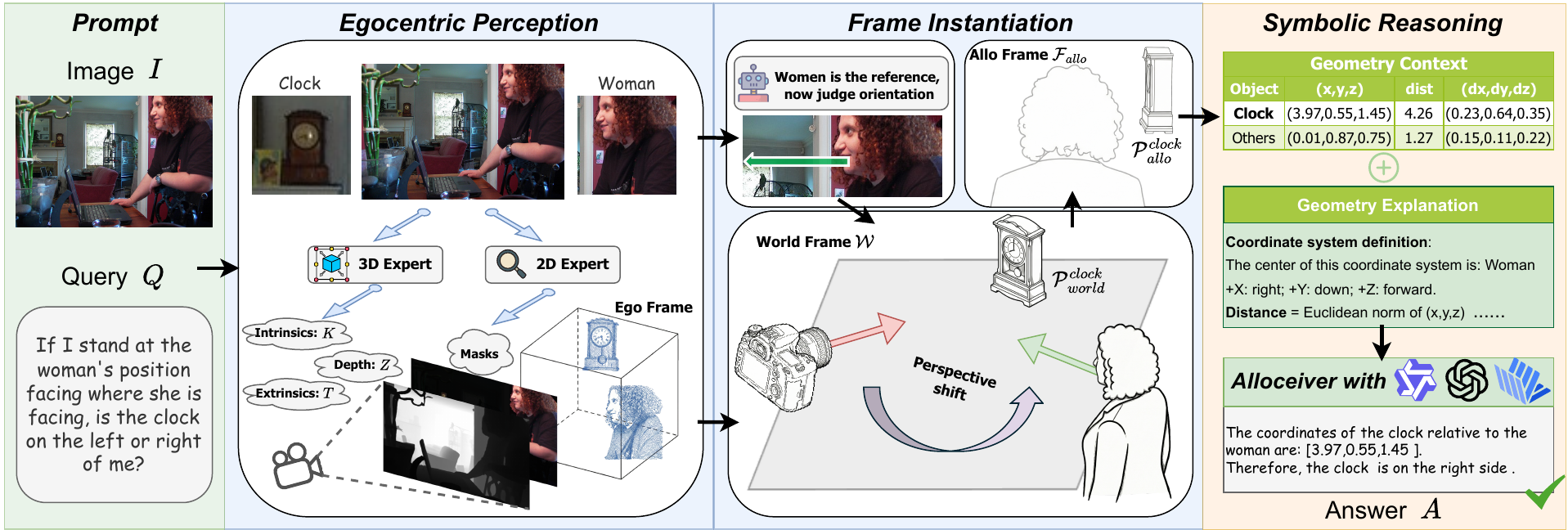}
  \caption{Framework of \textbf{Alloceiver}. To bridge the Reference Frame Gap, our framework explicitly decouples spatial reasoning from egocentric visual priors. The pipeline operates in three stages: (1) Metric-Aware Perception lifts 2D visual observations into a unified 3D metric world space ($\mathcal{W}$); (2) Dynamic Frame Instantiation constructs a query-aligned allocentric reference frame ($\mathcal{F}_{allo}$) via explicit coordinate transformation; and (3) Symbolic Geometry Reasoning derives the final answer through geometry-grounded logical deduction.}
  \label{framework}
\end{figure*}

\section{Method}
Formally, we define the task as deriving an answer $A$ given a visual scene set of $N$ images $\mathcal{I} = \{I_j\}_{j=1}^N$ (where $N \ge 1$) and a natural language query $Q$. Allocentric queries inherently necessitate a World-Referenced Geometric Frame ($\mathcal{F}_{world}$) independent of the observer's viewpoint.
We identify the fundamental challenge as bridging the \textbf{Reference Frame Gap}: the disconnect between the egocentric visual input and the required allocentric reasoning space. To bridge this gap, we explicitly recover 3D metric states in a unified world space $\mathcal{W}$ and construct a transformation $\mathcal{T}: \mathcal{W} \rightarrow \mathcal{F}_{allo}$ that aligns with the semantic intent of $Q$.
To address this, we propose the Allocentric Perceiver, a framework designed to decouple spatial understanding from egocentric visual priors. As illustrated in Fig. \ref{framework}, our approach operates in three distinct stages: (1) Metric-Aware Egocentric Perception, which lifts 2D visual observations into a robust 3D metric point cloud; (2) Dynamic Frame Instantiation, which transforms these coordinates into a target object centric reference frame based on the instruction's intent; and (3) Symbolic Geometry Reasoning, which utilizes structured prompts to derive the final answer via geometry-grounded logic.


\subsection{Metric-Aware Egocentric Perception \label{Sec3_1}} 
The foundation of our approach is to accurately localize task-relevant objects and recover their relative metric 3D structures within the egocentric camera frame. This process addresses two primary challenges: 1) grounding complex textual descriptions in 2D images and 2) resolving geometric ambiguities in 3D space.

\textbf{Coarse-to-Fine Semantic Grounding.} Given a user query $Q$, we first use the language part of VLM to extract a set of key object descriptions $\mathcal{D} = \{d_i\}_{i=1}^M$. To localize these objects, we employ LangSAM \cite{langsam}. However, open-vocabulary detectors often struggle with lengthy, attribute-heavy descriptions (e.g., ``the light-colored wooden picnic table"), leading to low recall. To address this, we implement a Hierarchical Semantic Relaxation mechanism. Specifically, in each image $I_j \in \mathcal{I}$, for a complex description $d_i$, we iteratively simplify it into a coarser prompt $d_{coarse}$ (e.g., ``picnic table ") to query the detector, generating a set of candidate proposals. We then crop these candidate regions and compute their Image-Text Matching (ITM) scores against the original detailed description $d_i$ using CLIP \cite{radford2021learningtransferablevisualmodels}. The candidate with the highest similarity score is selected and verified by the VLM. This process ensures unique and accurate 2D anchoring, yielding a segmentation mask $m_i^j$ for each object in each image.

\textbf{3D Lifting with Geometric Consensus.} To transition from 2D pixels to 3D space, we utilize the off-the-shelf 3D estimation model Depth-Anything-3 \cite{lin2025depth3recoveringvisual} to estimate the depth map $Z_j$ and camera intrinsics $K_j$ and extrinsics $T_j \in SE(3)$ of each image. This allows us to lift observations from all images into a unified Canonical World Coordinate System $\mathcal{W}$. Since depth estimation at object boundaries is often noisy, we apply morphological erosion to the mask $m_i^k$ before back-projection. For an object $i$ observed in view $j$, we back-project its eroded mask pixels $(u, v)$ into the local camera frame and apply the extrinsic transformation to obtain its world coordinates $\mathbf{p}_{world}^{i,j}$. These back-projected points across all $N$ views are aggregated as a point cloud $\mathbf{p}_{world}^{i}$ of the object $i$.
\begin{equation}
\mathbf{p}_{world}^{i,j} = T_j^{-1} \cdot \left( Z_{u,v} \cdot K_j^{-1} [u, v, 1]^T \right)
\end{equation}
\begin{equation}
\mathbf{p}_{world}^{i} = \bigcup_{j=1}^{N}\mathbf{p}_{world}^{i,j} 
\end{equation}
In multi-view scenarios, a single semantic concept might correspond to fragmented instances or outliers across different frames. To resolve this 3D ambiguity, we apply density-based clustering (DBSCAN) to group $\mathbf{p}_{world}^{i}$ into distinct spatial clusters. Then, we select the cluster with the highest average ITM score (computed from its corresponding 2D crops), forming the $\{\mathcal{P}_i\}_{i=1}^M$ for each object $i$ in the unified space $\mathcal{W}$. We define the robust centroid of $\mathcal{P}_i$: $\mathbf{c}_i \in \mathcal{W}$ as the location of object $i$, ready for coordinate transformation. 
 More details are provided in the \textbf{Appendix \ref{A.1}, \ref{A.2}}.

\subsection{Dynamic Frame Instantiation \label{Sec3_2}}

Having recovered the 3D states of objects in the world space $\mathcal{W}$, the second stage aims to construct a \textbf{Allocentric Reference Frame ($\mathcal{F}_{allo}$)} aligned with the query's semantic intent. The prerequisite for defining a local frame is identifying the physical anchor that serves as the ``observer." We parse the spatial dependency in the query $Q$ and extract the Reference Object $O_{ref}$. For instance, given the query \textit{``Where is the bag relative to the man?"}, the \textit{man} is designated as $O_{ref}$. The robust centroid of this object, denoted as $\mathbf{c}_{ref} \in \mathcal{W}$.

\textbf{Allocentric Frame Definition.}
Formally, the allocentric frame is defined by a transformation tuple $\mathcal{T} = \{\mathbf{O}, \mathbf{R}\}$, where $\mathbf{O} \in \mathbb{R}^3$ is the origin and $\mathbf{R} = [{v}_{right}, {v}_{down}, {v}_{front}] \in SO(3)$ is the rotation matrix constructed from the $O_{ref}$ reference frame's orthonormal basis vectors. This coordinate system follows OpenCV's right-hand rule. We strictly set the origin to the reference object's centroid ($\mathbf{O} = \mathbf{c}_{ref}$). However, the instantiation of the rotation matrix hinges on defining the canonical forward axis ${v}_{front}$, which is highly context-dependent and determined by either intrinsic semantic clues or extrinsic relational clues.

\textbf{Type I: Basis via Intrinsic Semantic Orientation.} When the query implies the reference object's inherent perspective (e.g., \textit{``from the man's perspective"}), the forward axis is governed by $O_{ref}$'s visual appearance, which can only be defined in a single image (camera frame). We employ an orientation expert as an aid to estimate the front orientation vector ${v}_{cam}$ within the local camera frame of $I_{j^*}$. Specifically, we use the masked visual crop of $O_{ref}$ to judge its orientation, aiming to reduce visual redundancy and interference from image positional bias. (a). For $O_{ref}$ as humans, we utilize a fine-grained Head Pose Estimator \cite{10477888} to capture gaze direction. (b). For other objects (e.g., animals, chairs), we implement a prompt ensemble voting strategy to obtain a robust consensus on their front." We then lift this local vector into the world space using the camera extrinsics $T_{j^*}$: ${v}_{front} = T_{j^*}^{-1} \cdot {v}_{cam}$. This operation lifts the 2D semantic appearance to 3D perception for subsequent geometric reasoning.

\textbf{Type II: Basis via External Geometric Constraints.} Unlike intrinsic poses which are local to an object, spatial reasoning in large-scale or unstructured environments often necessitates reference frames defined by extrinsic relational constraints. For instance, to resolve navigational descriptions (e.g., \textit{``standing at the table ($O_{ref}$) facing the door ($O_{aux}$)"}), we determine the forward axis by the geometric vector connecting the reference object $O_{ref}$ and the auxiliary object $O_{aux}$. Leveraging the robust centroids of the reference object ($\mathbf{c}_{ref}$) and the auxiliary object ($\mathbf{c}_{aux}$), we define the forward vector directly in the world space $\mathcal{W}$:
\begin{equation}
{v}_{front} = Normalize(\mathbf{c}_{aux} - \mathbf{c}_{ref})
\end{equation}

This geometric definition is inherently robust to occlusion or single-view ambiguity, as the centroids are aggregated from the geometric consensus of all observable frames. 

\textbf{Coordinate System Orthogonalization.}
Once the forward axis ${v}_{front}$ is instantiated, we construct the full orthonormal basis for $\mathcal{F}_{allo}$. ${v}_{down}$ is selected as the positive y-axis in the camera's perspective, which is the default setting to avoid confusion in perspective interpretation \cite{lin2025depth3recoveringvisual}. We compute ${v}_{right} = {v}_{down} \times {v}_{front} $, consequently, the rotation matrix $\mathbf{R}$ is constructed by stacking these normalized basis vectors column-wise.
Finally, all relevant scene entities $\{\mathcal{P}_i\}$ are transformed from the world space $\mathcal{W}$ to the local allocentric frame $\mathcal{F}_{allo}$:
\begin{equation}
\mathcal{P}_{allo}^{i} = \mathbf{R}^T (\mathcal{P}_i - \mathbf{O})
\end{equation}
This transformation effectively disentangles spatial reasoning from the observer's viewpoint, providing a canonical representation for symbolic reasoning. More details are in the \textbf{Appendix \ref{Prompts_orientation}}.

\subsection{Symbolic Geometry Reasoning \label{Sec3_3}}

As demonstrated in our feasibility study in Section \ref{Intro}, egocentric visual priors actively interfere with allocentric reasoning. To circumvent this, the final stage of our framework completely decouples visual inputs from the reasoning process, solely prompting the VLM with the clean, mathematically rigorous spatial states computed in Section \ref{Sec3_2}
\textbf{Structured Geometry-to-Language Prompting.} We generate a structured geometric context: $S_{geo}$ that translates the transformed allocentric coordinates $\mathcal{P}_{allo}$ into a standardized textual representation. First, we define: The coordinate system is centered on the reference object. The positive X-axis represents [Right], the positive Y-axis represents [Down], and the positive Z-axis represents [Forward]. Then we mainly provide the following information with their respective geometric meaning for all relevant objects other than the reference object: a) Robust center coordinates, b) Distance to the reference object, c) Dimensions along the x, y, and z axes. 

\textbf{Reasoning and Inference.} Finally, we construct the composite input $\mathcal{Q}_{final} = [S_{geo}, Q, Q_{rea}]$ by augmenting the geometric context and the query with a reasoning instruction $Q_{think}$, which enforces a chain-of-thought process for text-based logical deduction. The VLM, now operating in a text-only mode, utilizes its chain-of-thought capabilities to perform arithmetic and logical deduction on the geometric information. For example, to answer \textit{``Is the bag behind the man?"}, the model simply checks if the bag's $z$-coordinate in the man's frame is negative (assuming $+z$ is forward), yielding a robust, unambiguous answer $a$. The full prompt templates are provided in the \textbf{Appendix \ref{Prompts_Geometry_Reasoning}}.

\begin{figure*}[t]
  \centering
  \includegraphics[width=\textwidth]{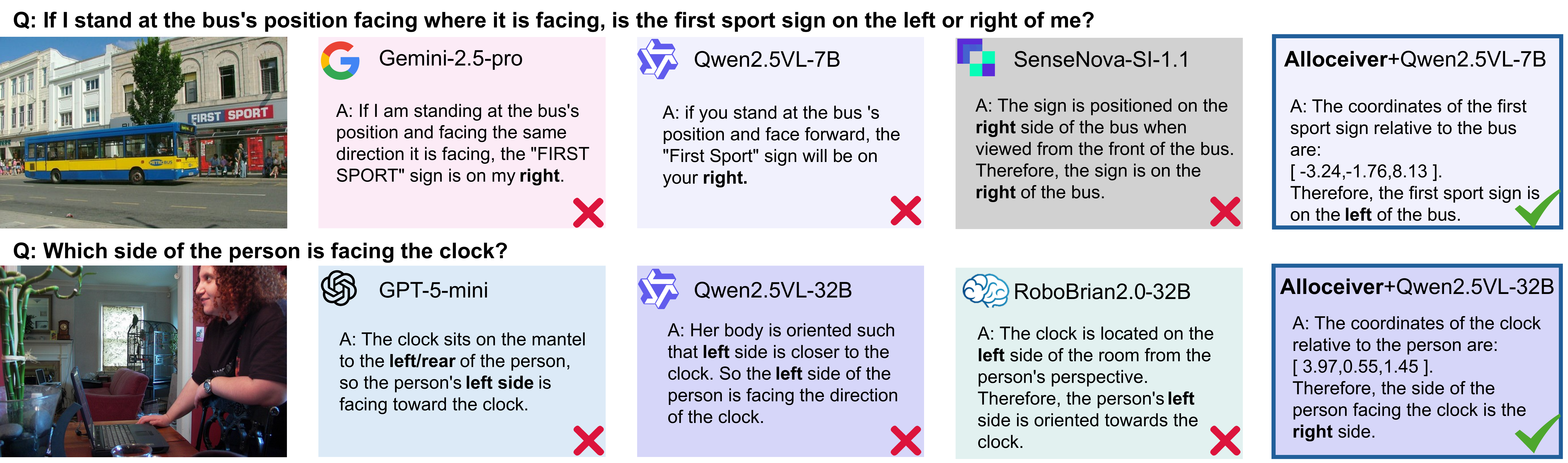}
  \caption{Comparison of Alloceiver's performance with other VLMs on typical allocentric questions. Although these VLMs (state-of-the-art commercial closed-source models and training models) attempt to substitute perspective, they are still affected by visual bias and reference frame gap, leading to failure.}
  \label{Cases}
\end{figure*}

\section{Experiments}

\subsection{Experimental Settings}
In the following sections, we denote our proposed framework as \textbf{Alloceiver} (Allocentric Perceiver).
We implement a \textbf{Spatial Query Router}. Designed as a training-free, plug-and-play module, our framework employs a lightweight semantic router to categorize queries into two streams. \textbf{Allocentric Stream}: Triggers the full dynamic frame instantiation (Sec. \ref{Sec3_2}) to resolve perspective shifts. \textbf{Egocentric Stream}: Utilizes the same Metric-Aware Perception (Sec. \ref{Sec3_1}) but retains the canonical camera frame, generating structured geometric prompts with the same 3D geometric cues (Sec. \ref{Sec3_3}) to enhance standard visual reasoning.

\textbf{Benchmarks.}
We evaluate our method on two public benchmarks focused on multi-perspective spatial reasoning:
(i) \textbf{ViewSpatial-Bench} \cite{li2025viewspatialbenchevaluatingmultiperspectivespatial}: Probes camera- and person-perspective queries, including relative direction, object-view orientation, and scene-simulation viewpoint shifts.
(ii) \textbf{3DSRBench} \cite{ma20253dsrbenchcomprehensive3dspatial}: We focus on viewpoint-conditioned tasks (\textit{orientation on the left, in front of, towards object}) and standard camera-view relations (\textit{location closer to object/camera}) to ensure robust performance across both allocentric and egocentric settings. Detailed descriptions of all evaluated tasks, including \textit{Orient. Left}, \textit{Orient. Twd.}, and others, along with illustrative samples, are provided in Appendix \ref{task_cases}.

\textbf{Models and baselines.}
Our primary setting evaluates the VLM as a base reasoner with our method as a training-free plug-in. To assess cross-backbone portability, we utilize heterogeneous models: Qwen2.5-VL (7B/32B), InternVL2.5-38B, and GPT-4o. 
For wider comparison, we include: 
(i) \textbf{Vanilla VLMs}: Both proprietary and open-source models, including GPT-5-mini \cite{openai_gpt5mini_docs}, Gemini-2.5-Pro \cite{comanici2025gemini25pushingfrontier}, GPT-4o \cite{openai_gpt4o_systemcard}, and InternVL2.5 \cite{chen2025expandingperformanceboundariesopensource}. 
(ii) \textbf{Spatially-tuned VLMs}: Models explicitly trained for spatial reasoning, such as Spatial-SSRL-7B \cite{liu2025spatialssrlenhancingspatialunderstanding}, Cosmos-Reason1-7B \cite{nvidia2025cosmosreason1physicalcommonsense}, SenseNova-SI-1.1 \cite{cai2025scalingspatialintelligencemultimodal}, SpatialThinker-7B \cite{batra2025spatialthinkerreinforcing3dreasoning}, REVPT-7B \cite{zhou2025reinforcedvisualperceptiontools}, and RoboBrain2.0-32B \cite{baairobobrainteam2025robobrain20technicalreport}. 
A \textbf{Random} baseline represents the expected accuracy of random guessing in multiple-choice tasks.

\textbf{Evaluation protocol.}
Both ViewSpatial-Bench and 3DSRBench are multiple-choice in our evaluated subsets. For all experiments, we use the official validation splits as the ground truth. Following standard protocols, we report Accuracy based on exact matching between the detected and the ground truth. Details regarding the router are provided in \textbf{Appendix \ref{Prompts_Router}}, and more implementation details are in \textbf{Appendix \ref{Implementation_details}}.

\newcommand{\inc}[1]{\ensuremath{^{ \textbf{\textcolor{mygreen}{+#1}}}}}
\newcommand{\dec}[1]{\ensuremath{^{ \textbf{\textcolor{myred}{-#1}}}}}
\newcommand{\graytext}[1]{\ensuremath{_{ \textcolor{gray}{#1}}}}

\begin{table*}[t]
\centering
\setlength{\tabcolsep}{2.5pt} 
\renewcommand{\arraystretch}{1.18}
\scriptsize
\caption{Results of Alloceiver enhancing different backbone VLM models on various egocentric or allocentric tasks. Tasks marked with superscript ${V}$ originate from the Viewspatial-Bench, and tasks marked with superscript ${T}$ are from the 3DSRBench. The metrics are expressed as percentage accuracy.}
\label{table1}
\begin{adjustbox}{width=\textwidth}
\begin{tabular}{@{}l*{7}{c}@{\hspace{6pt}}*{7}{c}@{}}
\toprule
\multicolumn{1}{c}{\multirow[c]{4}{*}{Tasks}}  & \multicolumn{7}{c}{Allocentric} & \multicolumn{7}{c}{Egocentric} \\
\cmidrule(lr){2-8}\cmidrule(lr){9-15}
 & \makecell{Psn.$^{V}$\\Orient.} & \makecell{Psn.$^{V}$\\Rel-Dir.} & \makecell{Psn.$^{V}$\\\textit{Sim-Dir.}} & \makecell{Orient.$^{T}$\\Front.} & \makecell{Orient.$^{T}$\\Left.} & \makecell{Orient.$^{T}$\\Twd.} & \makecell{Avg} 
 & \makecell{Cam.$^{V}$\\Orient.} & \makecell{Cam.$^{V}$\\Rel-Dir.} & \makecell{Orient.$^{T}$\\View.} & \makecell{Loc.$^{T}$\\Obj.} & \makecell{Loc.$^{T}$\\Cam.} & \makecell{Loc.$^{T}$\\Next.} & \makecell{Avg} \\
\midrule

Random      & 25.00 & 25.00 & 25.00 & 50.00 & 50.00 & 25.00 & 34.03 & 25.00 & 25.00 & 25.00 & 50.00 & 50.00 & 50.00 & 34.94  \\
\midrule
Qwen2.5VL-7B      & 36.04 & 35.51 & 26.15 & 59.45 & 35.24 & 20.04 & 32.92 & 27.81 & 45.40 & 42.86 & 58.71 & 81.49 & 60.03 & 52.26 \\
\rowcolor{gray!15}
\textbf{+Alloceiver}  & 42.36 & 48.81 & 28.87 & 57.70 & 51.29 & 32.43 & 41.25\inc{8.33} & 35.14  & 42.92 & 46.50 & 62.14 & 82.30 & 65.63 & 54.47\inc{2.21} \\
\midrule

Qwen2.5VL-32B     & 35.64 & 38.01 & 27.60 & 60.1 & 38.83 &  27.77 & 35.89 & 29.92 & 46.02 & 43.44 & 66.71 & 79.87 & 72.42 & 54.56 \\
\rowcolor{gray!15}
\textbf{+Alloceiver}  & 44.79 & 54.39 & 34.03 & 67.73 & 63.32 & 34.69 & 46.73\inc{10.84} & 57.43 & 54.03 & 44.24 & 69.29 & 79.87 & 79.35 & 61.72\inc{7.16} \\
\midrule

InternVL2.5-38B     &45.58	&42.76	&32.67	&61.92	&36.68	&29.15	&39.59 & 34.94	 & 53.98	& 47.08 &65.00	&83.04	&76.40	&58.91 \\
\rowcolor{gray!15}
\textbf{+Alloceiver}  & 47.62 & 57.24 & 39.91 & 67.15 & 66.76 & 40.09 & 50.32\inc{10.73} & 54.52 & 55.89 & 50.29 & 60.57 & 84.81 & 80.53 & 63.18\inc{4.27} \\
\midrule

GPT-4o    	& 43.78	& 45.01	& 29.14	& 58.43	& 44.70	& 30.54	& 39.81  & 19.78 & 50.48	& 47.16	& 67.86	& 79.72	& 74.48	& 52.27 \\
\rowcolor{gray!15}
\textbf{+Alloceiver}  & 47.42& 52.02 & 37.37 & 68.75 & 66.33 & 46.14 & 50.79\inc{10.98} & 26.41 & 58.88 & 56.71 & 71.29 & 82.45 & 75.37 & 60.55\inc{8.28} \\

\bottomrule
\end{tabular}
\end{adjustbox}
\end{table*}

\subsection{Enhancement Across VLMs}

In this section, we investigate the universality of Alloceiver as a training-free enhancement strategy. We report performance on both egocentric (which test fundamental 3D perception) and allocentric tasks (which require perspective shifts), as summarized in Tab. \ref{table1}. The results reveal that standard models struggle significantly with allocentric challenges. Notably, the average accuracy of Qwen2.5-VL-7B falls below the random baseline, while Qwen2.5-VL-32B performs only marginally better than chance. On particularly challenging subsets, such as \textit{Orient. Left} and \textit{Orient. Twd.}, performance frequently stagnates or drops below random-guess levels.

Our Alloceiver yields uniformly substantial improvements in allocentric performance across all backbones, boosting average accuracy by +8.33\% (Qwen2.5-VL-7B), +10.84\% (Qwen2.5-VL-32B), +10.73\% (InternVL2.5-38B), and +10.98\% (GPT-4o). The gains are most pronounced in tasks requiring strict orientation alignment, such as \textit{Psn. Rel-Dir}. (from the boy's perspective), \textit{Orient. Front}. and \textit{Orient. Left}. For instance, \textit{Orient. Left}. rises sharply under Alloceiver for every backbone (e.g., +16.05\% on Qwen2.5-VL-7B and +30.08\% on InternVL2.5-38B), indicating that adopting the allocentric frame (rather than expecting the backbone to internally simulate rotation) is the dominant driver. 

Crucially, our smallest augmented model, Qwen2.5-VL-7B + Alloceiver, significantly outperforms the unaugmented GPT-4o (39.81\%) on allocentric tasks. This validates a critical hypothesis: \textbf{\textit{Resolving allocentric queries necessitates intrinsic coordinate system transformations and geometric reasoning, rather than superficial visual-semantic pattern matching.}} Merely scaling up model parameters fails to address this issue, as the inherent visual-semantic ambiguity between egocentric and allocentric perspectives remains pervasive in training data describing relative positions and orientations. Consequently, for VLMs, the explicit Dynamic Frame Instantiation (Sec. \ref{Sec3_2}) is indispensable to decouple spatial logic from egocentric visual priors.



Alloceiver's motivation stems from allocentric ambiguity, yet it simultaneously boosts egocentric accuracy by 2.21\%--8.28\% across backbones. This demonstrates that structured 3D prompting do not trade off camera-view reasoning. While standard VLMs often falter on depth-sensitive queries (e.g., \textit{Loc. Obj.} or \textit{Loc. Next.}) due to their reliance on 2D monocular cues \cite{liao2024reasoningpathsreferenceobjects}, Alloceiver resolves this by lifting scenes into metric point clouds to provide precise distance constraints. The most significant gains occur in subtasks requiring stable 3D interpretation (\textit{Cam. Orient.}), reinforcing the premise that robust egocentric perception (Sec. \ref{Sec3_1}) is a prerequisite for allocentric reasoning.

Ultimately, these results establish Alloceiver as a portable, training-free enhancement that consistently elevates both allocentric and egocentric performance across open and proprietary VLMs.



\subsection{Comparison With More VLMs}

\begin{table*}[t]
\centering
\setlength{\tabcolsep}{2.5pt} 
\renewcommand{\arraystretch}{1.0}
\scriptsize
\caption{Performance over Viewspatial-Bench and 3DSRBench, across various model types. Top-1 \& Top-2 accuracies are represented using \textbf{bold text}, and \underline{underlines}. ``SenseNova-SI-1.1-Qwen2.5-VL-7B" model is abbreviated as ``SenseNova-SI-1.1-7B"}
\label{table2}
\begin{adjustbox}{width=\textwidth}
\begin{tabular}{@{}l*{7}{c}@{\hspace{6pt}}*{7}{c}@{}}
\toprule
\multicolumn{1}{c}{\multirow[c]{4}{*}{Tasks}}  & \multicolumn{7}{c}{Allocentric} & \multicolumn{7}{c}{Egocentric} \\
\cmidrule(lr){2-8}\cmidrule(lr){9-15}
 & \makecell{Psn.$^{V}$\\Orient.} & \makecell{Psn.$^{V}$\\Rel-Dir.} & \makecell{Psn.$^{V}$\\\textit{Sim-Dir.}} & \makecell{Orient.$^{T}$\\Front.} & \makecell{Orient.$^{T}$\\Left.} & \makecell{Orient.$^{T}$\\Twd.} & \makecell{Avg} 
 & \makecell{Cam.$^{V}$\\Orient.} & \makecell{Cam.$^{V}$\\Rel-Dir.} & \makecell{Orient.$^{T}$\\View.} & \makecell{Loc.$^{T}$\\Obj.} & \makecell{Loc.$^{T}$\\Cam.} & \makecell{Loc.$^{T}$\\Next.} & \makecell{Avg} \\
\midrule

Random      & 25.00 & 25.00 & 25.00 & 50.00 & 50.00 & 25.00 & 34.03 & 25.00 & 25.00 & 25.00 & 50.00 & 50.00 & 50.00 & 34.94  \\
\midrule
\rowcolor{gray!15}
\multicolumn{15}{c}{\textbf{\textit{Vanilla VLMs}}} \\

InternVL2.5-38B     &45.58	&42.76	&32.67	&61.92	&36.68	&29.15	&39.59 & 34.94	 & 53.98	& 47.08 &65.00	&83.04	&76.40	&58.91 \\
GPT-4o    	& 43.78	& 45.01	& 29.14	& 58.43	& 44.70	& 30.54	& 39.81  & 19.78 & 50.48	& 47.16	& 67.86	& 79.72	& 74.48	& 52.27 \\
Gemini-2.5-pro	& 46.08	& 42.28	& 31.76	& \underline{74.13}	& 38.11	& 32.22	& 41.81	& 33.33	& \underline{57.64}	& 51.75	& \underline{75.14}	 & 82.82	& \textbf{83.92}	& \underline{62.28} \\
GPT-5-mini	& 44.38	& 50.48	& 32.49	& \textbf{81.25} & 46.42	& 37.76	& 46.08	& 27.01	& 57.53	& \underline{53.94}	& \textbf{75.43}	& \underline{85.25}	& 72.27	& 61.13 \\
\midrule

\rowcolor{gray!15}
\multicolumn{15}{c}{\textbf{\textit{Trained Spatial VLMs}}} \\

RoboBrain2.0-32B	& 31.22	& 37.65	& 26.88	& 60.17	& 38.54	& 27.92	& 34.92	& 36.65	& 48.90	& 45.12	& 74.29	& \textbf{85.77}	& \underline{80.68}	& 59.36 \\
Spatial-SSRL-7B	    & 37.41	& 42.07	& 25.16	& 57.41 & 33.95 & 21.72 & 33.94 & 32.33 & 45.40 & 41.18 & 65.00 & 84.00 & 65.04 & 54.20 \\
Cosmos-Reason1-7B	& 46.69 & 37.77 & 19.10 & 60.61 & 40.40 & 21.65 & 34.91 & 23.59 & 42.64 & 40.60 & 56.71 & 77.88 & 59.44 & 49.51 \\
SenseNova-SI-1.1-7B	& \textbf{62.35} & 42.40 & \textbf{52.85} & 56.10 & 34.81 & 22.96 & 43.95 & 20.38 & 45.69 & 30.25 & 65.86 & 81.64 & 53.69 & 48.87 \\
SpatialThinker-7B	& 42.07 & 35.87 & 27.60 & 58.14 & 34.81 & 21.65 & 34.49 & 30.52 & 44.56 & 42.71 & 59.00 & 81.93 & 56.78 & 52.20 \\
REVPT-7B  & 39.76 & 36.34 & 27.51 & 61.05 & 35.24 & 21.94 & 34.61 & 29.62 & 43.65 & 40.31 & 63.57 & 82.08 & 61.50 & 49.69 \\
\midrule

\rowcolor{gray!15}
\multicolumn{15}{c}{\textbf{\textit{Alloceiver}}} \\

$\textbf{+}$ Qwen2.5VL-7B  & 42.36 & 48.81 & 28.87 & 57.70 & 51.29 & 32.43 & 41.25 & 35.14  & 42.92 & 46.50 & 62.14 & 82.30 & 65.63 & 54.47 \\

$\textbf{+}$ Qwen2.5VL-32B & 44.79 & \underline{54.39} & 34.03 & 67.73 & 63.32 & 34.69 & 46.73 & \textbf{57.43} & 54.03 & 44.24 & 69.29 & 79.87 & 79.35 & 61.72 \\

$\textbf{+}$ InternVL2.5-38B   & \underline{47.62} & \textbf{57.24} & \underline{39.91} & 67.15 & \textbf{66.76} & \underline{40.09} & \underline{50.32} & \underline{54.52} & 55.89 & 50.29 & 60.57 & 84.81 & 80.53 & \textbf{63.18} \\

$\textbf{+}$ GPT-4o   & 47.42 & 52.02 & 37.37 & 68.75 & \underline{66.33} & \textbf{46.14} & \textbf{50.79} & 26.41 & \textbf{58.88} & \textbf{56.71} & 71.29 & 82.45 & 75.37 & 60.55 \\

\bottomrule
\end{tabular}
\end{adjustbox}
\end{table*}

We further benchmark Alloceiver-enhanced backbones against two additional model groups: (i) strong vanilla VLMs, including proprietary open-source models; and (ii) trained spatial VLMs that start from the same vanilla backbones as ours (Qwen2.5-VL-7B or Qwen2.5-VL-32B) but undergo task-driven post-training for spatial intelligence. Results are summarized in Tab. \ref{table2}, while Tab. \ref{table3} aggregates average gains/losses relative to each model’s vanilla backbone. Performance examples as shown in Fig. \ref{Cases}

Tab. \ref{table2} demonstrates that Alloceiver achieves the best performance across both reference frames. Specifically, Alloceiver+GPT-4o reaches the highest Allocentric Avg (50.79\%), while Alloceiver+InternVL2.5-38B leads in Egocentric Avg (63.18\%) and ranks second in Allocentric Avg (50.32\%). Notably, this improvement transcends simple backbone scaling: Alloceiver+Qwen2.5-VL-32B achieves 46.73\% (Allocentric) and 61.72\% (Egocentric) averages, rivaling or exceeding larger proprietary models like GPT-5-mini and Gemini-2.5-Pro. These results confirm that Alloceiver functions as a capability multiplier, enhancing 3D cue utilization and reference-frame grounding beyond the limits of raw model scale.

Despite similar random baselines, both vanilla and spatially-trained VLMs favor egocentric perception, likely due to the abundance of first-person pretraining data versus the scarcity of high-quality, 3D-grounded allocentric supervision. Tab. \ref{table3} further reveals a performance trade-off in specialized models: for instance, SenseNova-SI-1.1-7B excels in allocentric tasks but suffers egocentric degradation, while RoboBrain2.0-32B shows the opposite trend. These patterns align with their respective training emphases on either perspective-taking or embodied interaction. 

In contrast, Alloceiver avoids this zero-sum trade-off. By enforcing reference-frame consistency at inference time via query routing and geometric injection, it substantially boosts allocentric reasoning while preserving or even enhancing the egocentric substrate across diverse backbones.

\subsection{Discussions}

\subsubsection{Whether Image inputting}



To eliminate geometric bias from visual priors, we derive the final decision solely from our geometry-grounded prompts, excluding image inputs. To evaluate prompt efficacy, we compare this \textit{w/o image} baseline against a \textit{w/ image} setting, where the original image is re-introduced at the last-step reasoning. Results in Tab.~\ref{table4} show only minor, inconsistent fluctuations across subtasks, with the overall average varying by merely $\sim$1.5\%. This performance parity suggests our prompts effectively neutralize visual prior influence by internalizing necessary spatial cues. Furthermore, the text-only configuration offers superior computational efficiency, particularly for multi-image or high-resolution scenarios.

\begin{table}[t]
\centering
\setlength{\tabcolsep}{1.7pt} 
\scriptsize
\caption{Comparison of Alloceiver performance with and without image input in the final inference stage.}
\label{table4}
\begin{adjustbox}{width=\linewidth}
\begin{tabular}{@{}lcccccc@{}}
\toprule
\multirow{2}{*}{Model} & \multicolumn{2}{c}{Psn.$^{V}$} & \multicolumn{3}{c}{Orient.$^{T}$} & \multirow{2}{*}{Avg} \\
\cmidrule(lr){2-3} \cmidrule(lr){4-6}
& Rel-Dir. & Abs-Dir. & Front. & Left. & Twd. & \\
\midrule
Q-7B \textit{w/o image} & 51.43 & 28.87 & 57.56 & 52.58 & 34.99 & \textbf{42.40} \\
Q-7B \textit{w/ image} & 52.38 & 25.43 & 60.17 & 45.70 & 33.97 & \textbf{40.83} \\
Q-32B \textit{w/o image} & 53.92 & 34.03 & 67.73 & 61.60 & 33.67 & \textbf{46.50} \\
Q-32B \textit{w/ image} & 56.18 & 34.21 & 67.59 & 61.03 & 37.61 & \textbf{47.99} \\
\bottomrule
\end{tabular}
\end{adjustbox}

\end{table}

\subsubsection{Improving robust spatial reasoning}


Rethinking the results in Tab. \ref{table2}--\ref{table3}, we confront a fundamental question: \textit{\textbf{How can we empower VLMs with human-like spatial intelligence that consistently generalizes across diverse reference frames beyond their training distributions?}} A retrospective analysis suggests that existing data-driven paradigms may have hit a critical bottleneck. For instance, while SenseNova-SI \cite{cai2025scalingspatialintelligencemultimodal} exhibit proficiency in specific tasks such as \textit{Psn. Orient.}, this localized success fails to extrapolate to broader allocentric scenarios (e.g., \textit{Orient. Left/Twd.}) and often incurs a ``catastrophic forgetting" of egocentric competence. Such performance plateaus indicate that scaling taxonomy-driven datasets may not foster internalized, generalizable reasoning. In contrast, Spatial-SSRL eliminates such trade-offs by employing ego/allo self-supervision and post training with verifiable geometric signals. This highlights that explicit geometric verification, rather than raw data volume, is a more feasible path toward robust spatial intelligence.

Future training-based research should prioritize constructing datasets with explicit reference-frame annotations and multi-frame reasoning objectives to bridge the ego/allo gap. In this paradigm, Alloceiver can act as a teacher by generating geometrically grounded reasoning traces to supervise spatially consistent intelligence. Nevertheless, our training-free approach remains a highly effective and portable solution that bypasses the prohibitive costs of data curation and model retraining.

\begin{table}[t]
\centering
\setlength{\tabcolsep}{2.5pt} 
\renewcommand{\arraystretch}{1}
\scriptsize
\caption{Performance gain or loss of trained spatial VLMs and Alloceiver-enhanced VLMs compared to Vanilla VLMs.}
\label{table3}
\begin{adjustbox}{width=\linewidth}
\begin{tabular}{c c c}
\toprule
\multicolumn{1}{c}{Tasks} & Allocentric Avg & Egocentric Avg \\
\midrule
\rowcolor{gray!15}
Qwen2.5VL-7B    & 32.92  & 52.26 \\

Spatial-SSRL-7B	     & 33.94\inc{1.02}  & 54.20\inc{1.94} \\
Cosmos-Reason1-7B	 & 34.91\inc{1.99} & 49.51\dec{2.75} \\
SenseNova-SI-1.1-7B	 & 43.95\inc{11.03} & 48.87\dec{3.39} \\
SpatialThinker-7B	& 34.4\inc{1.48} & 52.20\dec{0.06} \\
REVPT-7B  & 34.61\inc{1.69} & 49.69\dec{2.57} \\
\textbf{Alloceiver$+$Qwen2.5VL-7B} & 41.25\inc{8.33}  & 54.47\inc{2.21} \\
\midrule

\rowcolor{gray!15}
Qwen2.5VL-32B    & 35.89 & 54.56 \\

RoboBrain2.0-32B & 34.92\dec{0.97} & 59.36\inc{4.8} \\

\textbf{Alloceiver$+$Qwen2.5VL-32B} & 46.73\inc{10.84}  & 61.72\inc{7.16} \\
\bottomrule
\end{tabular}
\end{adjustbox}
\end{table}



\section{Conclusion}

We presented Allocetric Perceiver, a training-free framework for robust allocentric spatial reasoning that decouples geometric inference from egocentric visual priors. By lifting observations into a unified metric world space, instantiating an instruction-aligned object-centric frame through explicit transformations, and prompting the backbone with structured geometry-to-language representations, Alloceiver reduces allocentric decision-making to verifiable, geometry-grounded deduction. Across two multi-perspective spatial benchmarks and diverse VLM backbones, Alloceiver consistently improves allocentric accuracy without sacrificing egocentric competence, demonstrating that reliable perspective-taking hinges on explicit reference-frame grounding and geometric constraints. 

\section*{Impact Statement}
This paper presents work whose goal is to advance the field of Machine Learning. There are many potential societal consequences of our work, none which we feel must be specifically highlighted here.

\nocite{langley00}

\bibliography{example_paper}
\bibliographystyle{icml2026}

\newpage
\appendix
\onecolumn

\section{Methods Details}

\subsection{Iterative Coarse-to-Fine Grounding} \label{A.1}

Algorithm \ref{alg:grounding_iterative} details our Iterative Semantic Relaxation mechanism. Unlike one-step simplification, this process (Lines 5-18) progressively peels away linguistic modifiers layer-by-layer (e.g., "light-colored wooden picnic table" $\rightarrow$ "wooden picnic table" $\rightarrow$ "picnic table"). The rationale is twofold: \textbf{Recall Maximization}: Open-vocabulary detectors often fail to recognize objects when prompts are overly specific. We iterate until the detector returns a sufficient pool of candidates ($|\mathcal{B}| > 1$), ensuring we do not miss the target due to strict textual constraints. \textbf{Safety Backtracking}: We implement a heuristic check (Lines 9-12) to handle edge cases where simplification inadvertently broadens the semantic scope to irrelevant categories or causes detector failure. If a simplified prompt yields fewer detections than its more specific predecessor, the algorithm terminates early and reverts to the previous, more effective result.Once a candidate pool is established, the Fine-Grained Selection phase (Lines 19-27) strictly enforces the original attributes by ranking candidates against the initial detailed query $d_{orig}$, ensuring the final output is both visually grounded and semantically faithful.

\begin{figure}[h]
  \centering
  \includegraphics[width=0.8\linewidth]{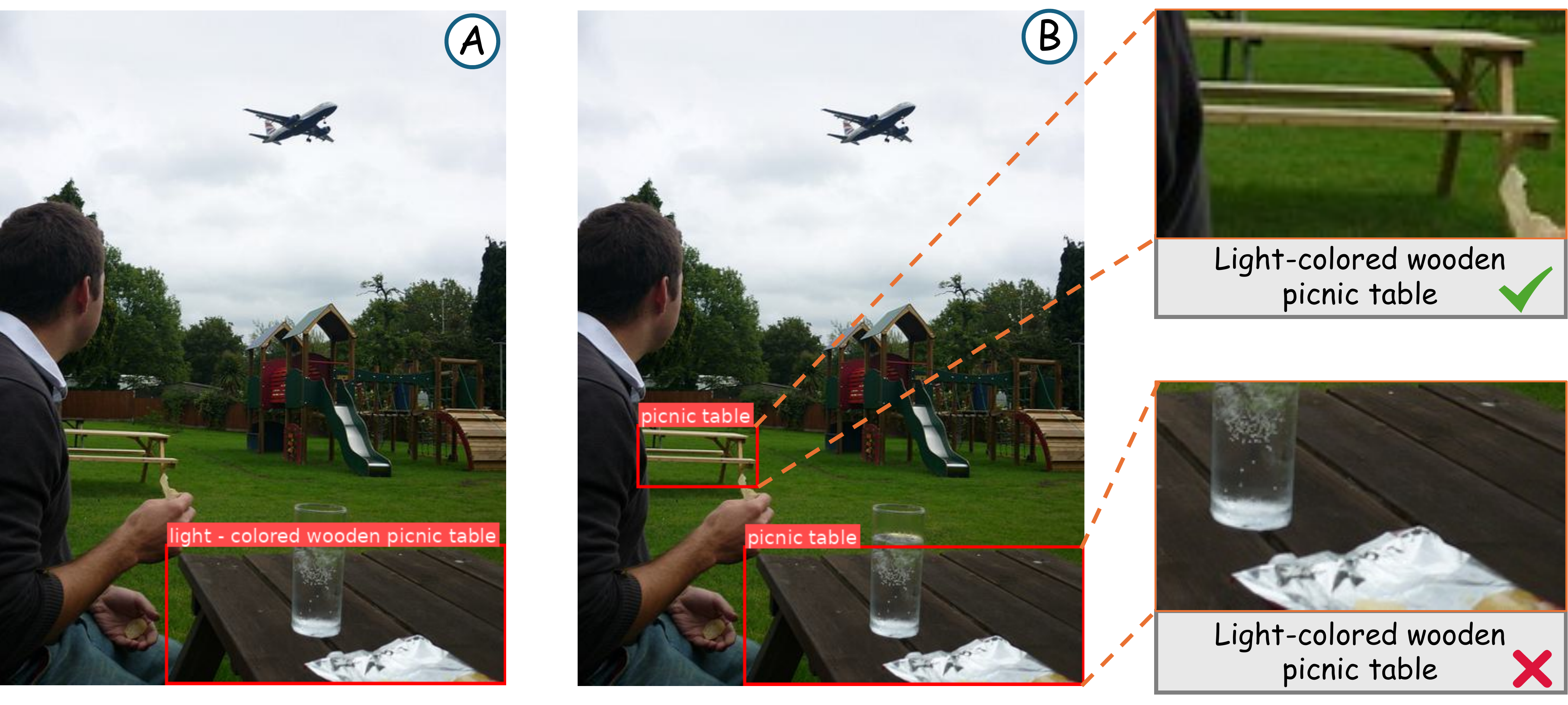}
  \caption{Examples concrete implementation of Coarse-to-Fine Semantic Grounding.}
  \label{Grouding}
\end{figure}

As illustrated in Fig. \ref{Grouding}, the query 'light-colored wooden picnic table' is erroneously grounded to a 'dark-colored' instance in panel (a). Following semantic relaxation, the simplified concept 'picnic table' retrieves two distinct instances in panel (b). By computing ITM scores between these candidates and the original fine-grained description, our approach ultimately achieves accurate visual-semantic grounding.


\noindent\rule{\textwidth}{1pt} 
\captionof{algorithm}{Iterative Coarse-to-Fine Semantic Grounding} 
\label{alg:grounding_iterative}
\noindent\rule{\textwidth}{0.5pt} 
\begin{algorithmic}[1]
   \STATE {\bfseries Input:} Image $I$; Detailed object description $d_{orig}$; VLM for text simplification $\text{VLM}_{degen}$; Open-Vocabulary Detector $\text{Det}$; Image-Text Matcher $\text{CLIP}$.
   \STATE {\bfseries Output:} Best matched object mask $m^*$.
   
   \STATE Initialize candidate set $\mathcal{B}_{final} \leftarrow \emptyset$
   \STATE $d_{curr} \leftarrow d_{orig}$ \COMMENT{Start with the most specific description}
   \STATE $\mathcal{R}_{history} \leftarrow []$ \COMMENT{Track detection history for backtracking}

   \STATE \textit{// Phase 1: Iterative Semantic Relaxation (Recall Maximization)}
   \WHILE{True}
       \STATE $\mathcal{B}_{curr} \leftarrow \text{Det}(I, d_{curr})$ \COMMENT{Detect with current prompt}
       \STATE Append $\mathcal{B}_{curr}$ to $\mathcal{R}_{history}$

       \STATE \textit{// Condition 1: Safety Backtracking}
       \IF{$|\mathcal{R}_{history}| > 1$ \AND $|\mathcal{B}_{curr}| < |\mathcal{R}_{history}[-2]|$}
           \STATE $\mathcal{B}_{final} \leftarrow \mathcal{R}_{history}[-2]$ \COMMENT{Revert if simplification hurts recall}
           \STATE \textbf{break}
       \ENDIF

       \STATE \textit{// Condition 2: Success or Convergence}
       \STATE $d_{next} \leftarrow \text{VLM}_{degen}(d_{curr})$ \COMMENT{Remove ONE layer of modifiers}
       \IF{$|\mathcal{B}_{curr}| > 1$ \OR $d_{next} == d_{curr}$}
           \STATE $\mathcal{B}_{final} \leftarrow \mathcal{B}_{curr}$
           \STATE \textbf{break} \COMMENT{Found candidates or cannot simplify further}
       \ENDIF

       \STATE $d_{curr} \leftarrow d_{next}$ \COMMENT{Continue with simplified prompt}
   \ENDWHILE

   \STATE \textit{// Phase 2: Fine-Grained Selection (Precision Alignment)}
   \STATE Initialize best score $s^* \leftarrow -1$, best mask $m^* \leftarrow \text{None}$
   
   \FOR{\textbf{each} box $b_k \in \mathcal{B}_{final}$}
       \STATE $I_{crop} \leftarrow \text{Crop}(I, b_k)$
       \STATE $s_{itm} \leftarrow \text{CLIP}(I_{crop}, d_{orig})$ \COMMENT{Score against ORIGINAL detailed text}
       \IF{$s_{itm} > s^*$}
           \STATE $s^* \leftarrow s_{itm}$
           \STATE $m^* \leftarrow \text{Mask}(b_k)$
       \ENDIF
   \ENDFOR

   \STATE \textbf{return} $m^*$
\end{algorithmic}
\noindent\rule{\textwidth}{1pt} 

\subsection{Implementation Details of 3D Lifting with Geometric Consensus} \label{A.2}

\begin{figure}[h]
  \centering
  \includegraphics[width=0.8\linewidth]{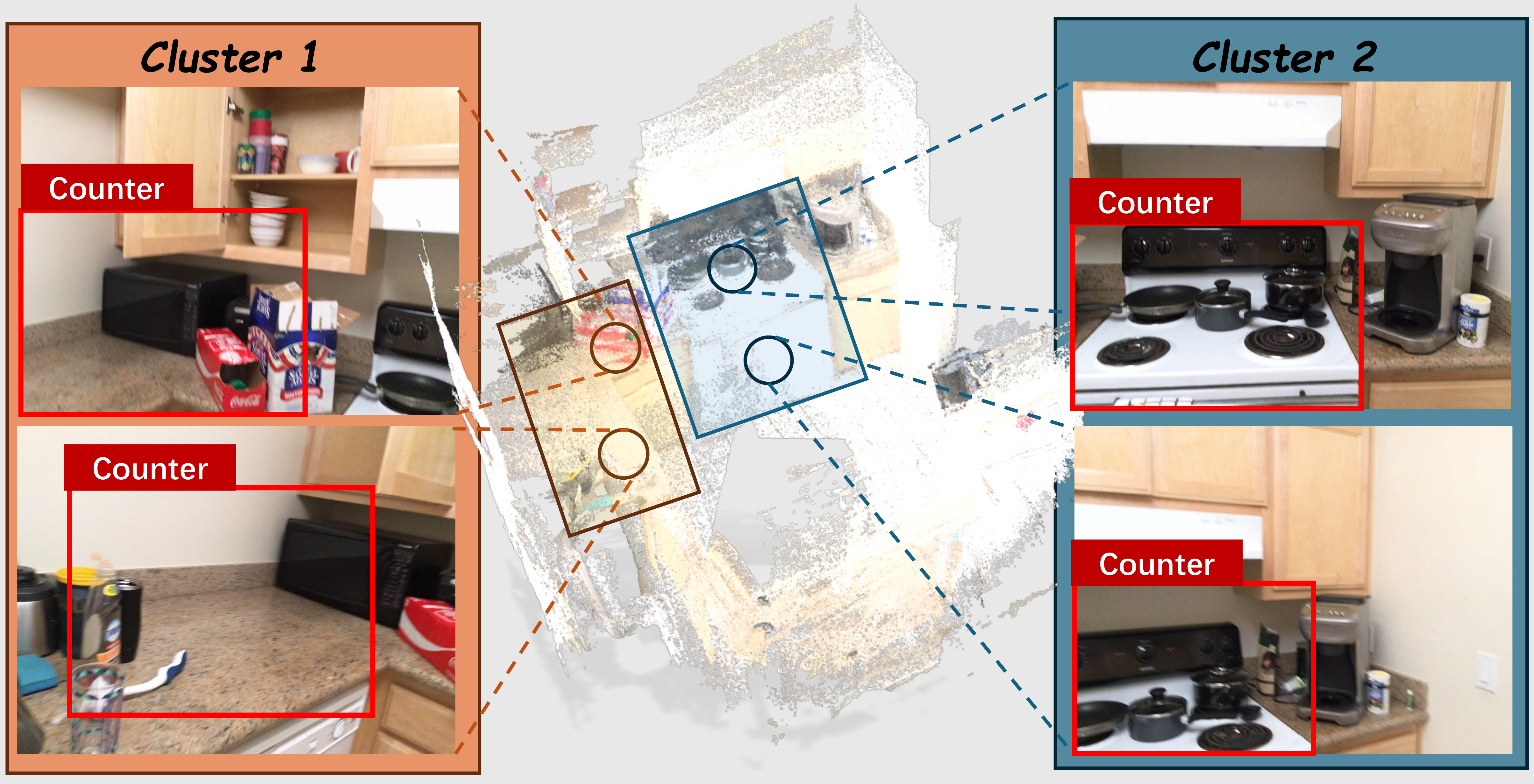}
  \caption{Schematic diagram of clusters in 3D space and their corresponding viewpoints images.}
  \label{DBSCAN}
\end{figure}

\textbf{Geometric consensus for 3D lifting.}
Algorithm~\ref{alg:geo_consensus_lifting} implements the geometric-consensus stage of our metric-aware egocentric perception pipeline.
Given a set of view-specific 2D instance hypotheses (masks/boxes) for an object $o$, we first lift each hypothesis to a partial 3D point cloud in the \textit{world} coordinate frame and then select a consistent subset via clustering and semantic ranking.

\textbf{Correct world-frame lifting.}
For a pixel $(u,v)$ inside an instance mask in view $j$, we obtain its depth $z=Z_j(u,v)$ and back-project it into camera coordinates as
$p_j^{cam}= z\,K_j^{-1}[u,v,1]^\top$.
We then transform it to the world frame using the camera pose.
$p_j^{world}= T_j^{-1} p_j^{cam}$.

\textbf{DBSCAN and ITM-ranked cluster selection.}
To resolve multi-view ambiguity (e.g., distinguishing multiple instances of a concept across frames), we perform DBSCAN clustering on lifted FPS points in world space. View-specific instances are assigned to clusters via majority voting. We then employ a semantic tie-breaker: the cluster maximizing the mean Image-Text Matching (ITM) score of its constituent crops is selected. This synergizes cross-view geometric consensus with semantic alignment, ensuring the final 3D representation (centroid and extent) is both spatially coherent and linguistically accurate.

For the Figure Caption (Fig. 7):
Figure \ref{DBSCAN}: Geometric Disambiguation. The query "Counter" initially anchors to distinct instances (a kitchen counter and a desk counter). DBSCAN segregates these into separate spatial clusters; we then select the cluster with the highest consensus ITM score as the canonical 3D representation for downstream reasoning.

\noindent\rule{\textwidth}{1pt}
\captionof{algorithm}{3D Lifting with Geometric Consensus (ITM-Ranked DBSCAN Selection)}
\label{alg:geo_consensus_lifting}
\noindent\rule{\textwidth}{0.5pt}
\begin{algorithmic}[1]
   \STATE {\bfseries Input:} Key objects $\mathcal{O}$; per-view detections $\{\texttt{info}_{o,j}\}$ (mask $m_{o,j}$, box $b_{o,j}$);
   depth maps $\{Z_j\}_{j=1}^N$; intrinsics $\{K_j\}_{j=1}^N$; extrinsics $\{T_j\}_{j=1}^N$ with $T_j\in SE(3)$ (camera $\to$ world, as in Eq.(1) in the main text);
   images $\{I_j\}_{j=1}^N$; ITM scorer $\text{ITM}(\cdot,\cdot)$; DBSCAN params $(\varepsilon, s_{\min})$; FPS budget $k$.
   \STATE {\bfseries Output:} For each $o\in\mathcal{O}$: world point cloud $P_o$; robust center $c_o$; extent $e_o$.

   \FOR{\textbf{each} object $o \in \mathcal{O}$}
      \STATE $\mathcal{S} \leftarrow [\ ]$ \COMMENT{view-specific 3D instance hypotheses}
      \FOR{\textbf{each} view $j \in \{1,\dots,N\}$}
         \IF{$\texttt{info}_{o,j}$ is invalid}
            \STATE \textbf{continue}
         \ENDIF
         \STATE $(m_{o,j}, b_{o,j}) \leftarrow \texttt{info}_{o,j}$

         \STATE $m'_{o,j} \leftarrow \text{Erode}(m_{o,j})$ \COMMENT{reduce boundary depth noise}
         \STATE $\Omega_{o,j} \leftarrow \{(u,v)\mid m'_{o,j}(u,v)=1\}$

         \STATE \textit{// Back-project masked pixels to camera coordinates}
         \STATE $P^{cam}_{o,j} \leftarrow [\ ]$
         \FOR{\textbf{each} $(u,v) \in \Omega_{o,j}$}
            \STATE $z \leftarrow Z_j(u,v)$
            \STATE $p^{cam} \leftarrow z \cdot K_j^{-1}[u,v,1]^\top \in \mathbb{R}^3$
            \STATE append $p^{cam}$ to $P^{cam}_{o,j}$
         \ENDFOR
         \STATE $P^{cam}_{o,j} \leftarrow \text{CleanOutliers}(P^{cam}_{o,j})$

         \STATE \textit{// Camera $\to$ World via extrinsics $T_j$ (consistent with Eq.(1))}
         \STATE $P^{world}_{o,j} \leftarrow \left\{\, \big(T_j^{-1} \cdot [p^{cam};1]\big)_{1:3} \ \middle|\ p^{cam}\in P^{cam}_{o,j} \,\right\}$

         \STATE $s_{o,j} \leftarrow \text{ITM}(\text{Crop}(I_j,b_{o,j}), \text{text}(o))$
         \STATE $P^{fps}_{o,j} \leftarrow \text{FPS}(P^{world}_{o,j}, k)$ \COMMENT{for efficient clustering}
         \STATE append instance $\left(P^{world}_{o,j}, P^{fps}_{o,j}, s_{o,j}\right)$ to $\mathcal{S}$
      \ENDFOR

      \STATE \textbf{assert} $|\mathcal{S}|>0$

      \STATE \textit{// DBSCAN on concatenated FPS points (point-level clustering in world space)}
      \STATE $P^{all} \leftarrow \bigcup\limits_{(P^{fps},\cdot,\cdot)\in\mathcal{S}} P^{fps}$
      \STATE $\ell(\cdot) \leftarrow \text{DBSCAN}(P^{all}; \varepsilon, s_{\min})$

      \STATE \textit{// Convert point clusters to instance clusters via majority vote}
      \FOR{\textbf{each} instance index $t$ in $\{1,\dots,|\mathcal{S}|\}$}
         \STATE $\mathcal{L}_t \leftarrow \{\ell(p)\mid p\in P^{fps}_t\}$
         \STATE $g_t \leftarrow \text{Mode}(\mathcal{L}_t)$
      \ENDFOR
      \STATE $\{\mathcal{C}_g\} \leftarrow$ group instances by $g_t$ \COMMENT{$\mathcal{C}_g$ is a set of instance indices}

      \STATE \textit{// Geometric consensus: pick the cluster with highest mean ITM (exclude DBSCAN noise label $-1$)}
      \STATE $g^\star \leftarrow \arg\max\limits_{g \neq -1}\ \frac{1}{|\mathcal{C}_g|}\sum\limits_{t\in\mathcal{C}_g} s_t$
      \IF{no valid $g^\star$ exists (e.g., all $g=-1$)}
         \STATE $t^\star \leftarrow \arg\max\limits_t\ s_t$; \ \ $\mathcal{C}^\star \leftarrow \{t^\star\}$
      \ELSE
         \STATE $\mathcal{C}^\star \leftarrow \mathcal{C}_{g^\star}$
      \ENDIF

      \STATE \textit{// Merge selected cluster to form final object geometry}
      \STATE $P_o \leftarrow \bigcup\limits_{t\in\mathcal{C}^\star} P^{world}_t$
      \STATE $c_o \leftarrow \text{RobustCentroid}(P_o)$
      \STATE $e_o \leftarrow \text{RobustExtent}(P_o)$
   \ENDFOR

   \STATE \textbf{return} $\{(P_o, c_o, e_o)\}_{o\in\mathcal{O}}$
\end{algorithmic}
\noindent\rule{\textwidth}{1pt}


\section{Experimental Details}

\subsection{Implementation details.} \label{Implementation_details}
For all backbone VLMs, we employ greedy decoding (temperature set to 0) with the random seed fixed at 42 to ensure reproducibility. Regarding hardware configuration, inference for 7B models is conducted on a single NVIDIA A6000 GPU, whereas 32B and 38B models are distributed across four NVIDIA A6000 GPUs. Proprietary commercial models are accessed via their respective official APIs. Additionally, all 2D and 3D experts are deployed on a single NVIDIA A6000 GPU.

\subsection{Task Cases} \label{task_cases}

Tasks marked with superscript ${V}$ originate from the Viewspatial-Bench:
camera perspective object View orientation (\textit{Cam. Orient.}),
camera perspective relative direction (\textit{Cam. Rel-Dir.}),
Person perspective Object view orientation (\textit{Psn. Orient.}), person perspective relative direction (\textit{Psn. Rel-Dir}.), and person perspective scene simulation relative direction (\textit{Sim-Dir.}). Tasks marked with superscript ${T}$ originate from the

3DSRBench: orientation viewpoint (\textit{Orient. View.}), location closer to camera (\textit{Loc. Cam.}), location closer to object (\textit{Loc. Obj.}), location next to (\textit{Loc. Next.}), orientation in front of (\textit{Orient. Front}.), orientation on the left (\textit{Orient. Left}.), viewpoint towards object (\textit{Orient. Twd.})

\begin{figure}[h]
  \centering
  \includegraphics[width=\linewidth]{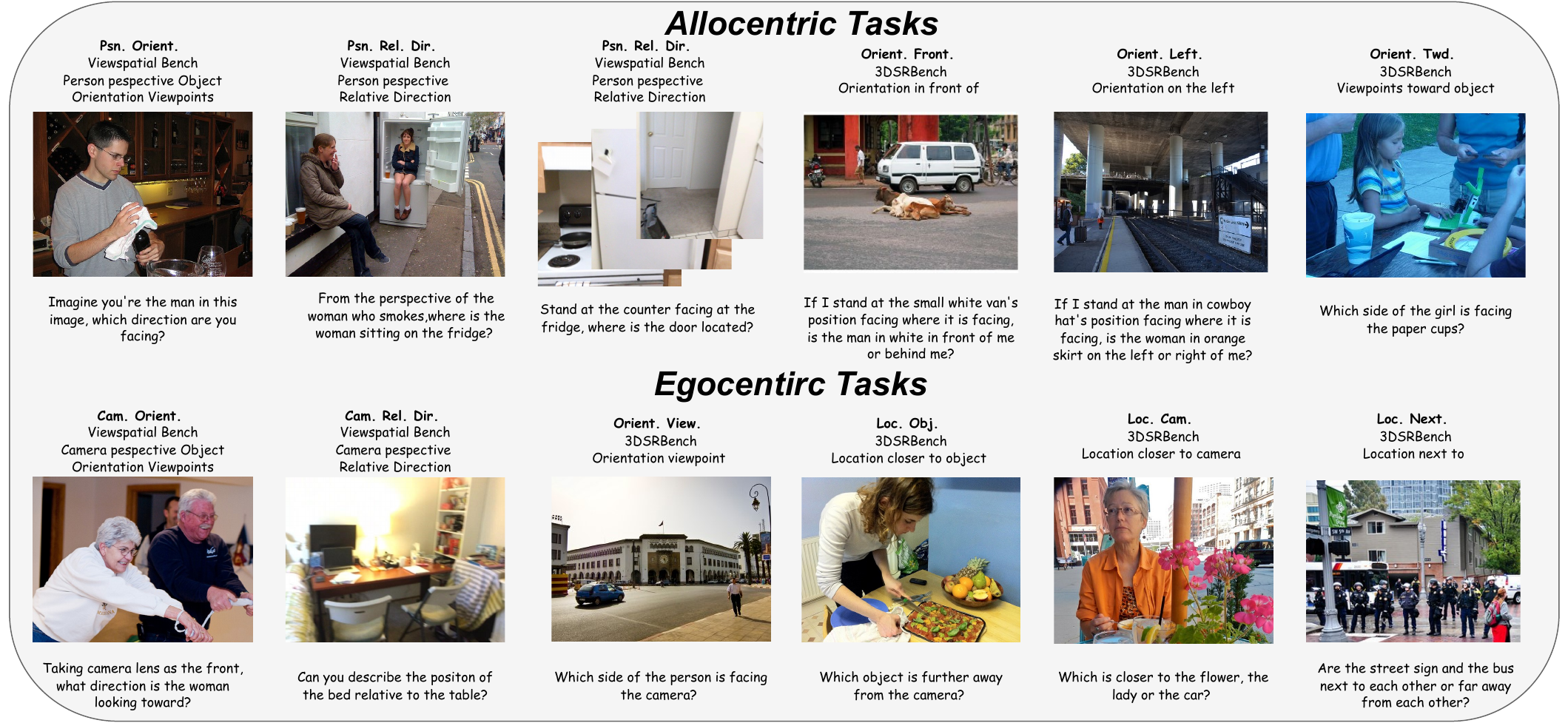}
  \caption{The specific category names of the tested subtasks in respective benchmarks and their corresponding question image-text examples.}
  \label{question_samples}
\end{figure}

Examples of all tasks are shown in the Fig. \ref{question_samples}

\section{Prompts for Allocentric Perceiver}

\subsection{Prompts for Key Object Extraction} 

Given the text query \(Q\), we input it into the following Prompt template to extract the description of all the key objects involved in the question answering.

\tcbset{
    vqabox/.style={
        breakable,
        colback=gray!10!white, 
        colframe=gray!50!black, 
        boxrule=0.5pt, 
        arc=2mm, 
        left=4pt, right=4pt, top=4pt, bottom=4pt, 
        fonttitle=\bfseries, 
        title style={fill=gray!30!white, coltext=black} 
    }
}


\begin{tcolorbox}[vqabox]
\small
\textbf{Prompt: Key object (core subject) extraction from the question.} 
\medskip
\noindent\textbf{Goal.} Extract all core subjects that directly participate in answering the question, while excluding personal pronouns.

\medskip
\noindent\textbf{Prompt template:}
\begin{quote}\ttfamily
Task: Extract all subjects that directly participate in the question's answer with their complete descriptions from the question. Do not extract any personal pronouns (such as I, you, he, she, it, etc.). For example, from 'Is the person with white trousers on the left or right side of the person in blue?', extract 'person with white trousers' and 'person in blue' (both are core subjects for the position relationship question).\\
Question: \{\(Q\)\}\\
Action: Only list the core subjects in the order they appear in the 'Question' separated by commas.
\end{quote}

\noindent\textbf{Expected output format.} A single comma-separated list, preserving the original descriptions and the order of appearance.
\end{tcolorbox}

\subsection{Prompts for VLM to Judge Orientation} \label{Prompts_orientation}

When determining the orientation of a target object, we input the object's name, \{object\_keyword\}, into Prompts A, B, and C below. After obtaining multiple answers from the VLM output, the final result will be selected by a vote among these answers.

\tcbset{
    vqabox/.style={
        breakable,
        colback=gray!10!white, 
        colframe=gray!50!black, 
        boxrule=0.5pt, 
        arc=2mm, 
        left=4pt, right=4pt, top=4pt, bottom=4pt, 
        fonttitle=\bfseries, 
        title style={fill=gray!30!white, coltext=black} 
    }
}


\begin{tcolorbox}[vqabox]

\small
\textbf{Prompt A: Single-shot 8-way orientation classification.}

\medskip
\noindent\textbf{Candidate label set (fixed):}

\begin{quote}\ttfamily
\{\texttt{front},\texttt{left-front},\texttt{left},\texttt{left-back},\texttt{back},\texttt{right-back},\texttt{right},\texttt{front-right}\}.
\end{quote}

\medskip
\noindent\textbf{Prompt template:}
\begin{quote}\ttfamily
Which direction is the \{object\_keyword\} facing in the image?\\
Choose from \{front, left-front, left, left-back, back, right-back, right, front-right\}.\\

\end{quote}

\noindent\textbf{Note:} This prompt directly asks for an 8-way direction label “in the image,” i.e., from the viewer/camera perspective.
\end{tcolorbox}

\begin{tcolorbox}[vqabox]
\small
\textbf{Prompt B: Three-round coarse-to-fine interrogation (Binary $\times$2 + final 3-way; \texttt{vlm\_orientation\_judge}).}

\medskip
\noindent\textbf{Round 1 (front vs. back):}
\begin{quote}\ttfamily
Is the \{object\_keyword\} facing forward or backward in the image from viewer's perspective?\\
Answer front or back in \textbackslash boxed\{\}.
\end{quote}

\noindent\textbf{Round 2 (left vs. right):}
\begin{quote}\ttfamily
Is the \{object\_keyword\} facing left or right in the image from viewer's perspective?\\
Answer left or right in \textbackslash boxed\{\}.
\end{quote}

\noindent\textbf{Round 3 (local refinement within a 3-way candidate set):}

\noindent The candidate set is deterministically constructed from the first two answers:
\[
\begin{array}{ll}
(\texttt{front},\texttt{right}) \mapsto \{\texttt{front},\texttt{front-right},\texttt{right}\}, &
(\texttt{front},\texttt{left}) \mapsto \{\texttt{front},\texttt{left-front},\texttt{left}\},\\
(\texttt{back},\texttt{right}) \mapsto \{\texttt{back},\texttt{right-back},\texttt{right}\}, &
(\texttt{back},\texttt{left}) \mapsto \{\texttt{back},\texttt{left-back},\texttt{left}\}.
\end{array}
\]

\noindent\textbf{Prompt template:}
\begin{quote}\ttfamily
Which direction is the \{object\_keyword\} facing in the image from viewer's perspective?\\
Choose from \{ \{options\_str\} \}.\\

\end{quote}

\noindent\textbf{Note:} This strategy reduces confusion by first resolving the coarse axis (front/back, left/right) and then disambiguating the diagonal vs. axis-aligned direction.
\end{tcolorbox}

\begin{tcolorbox}[vqabox]
\small
\textbf{Prompt C: Unified definition on orientation.}

\medskip
\noindent\textbf{Task definition about orientation}
\begin{quote}\ttfamily
You are an expert at judging the facing direction of an object relative to the CAMERA.\\
You will see a cropped image that mainly contains ONE target object.\\
Your task is to decide which direction the FRONT/FACING side of this object is pointing relative to the CAMERA.\\[2pt]
Rules:\\
1. All directions are defined from the CAMERA'S point of view using the IMAGE axes:\\
\ \ - "front": the object's front/face points toward the camera.\\
\ \ - "back": the object's front/face points away from the camera.\\
\ \ - "left" / "right": the object's front/face points to the left/right side of the IMAGE.\\
\ \ - "front-left", "front-right", "back-left", "back-right": between these main axes.\\
2. For PEOPLE or ANIMALS, use the orientation of the HEAD/FACE (nose, eyes, and mouth) to define the facing direction.\\
\ \ - If the head is turned relative to the body, ALWAYS follow the HEAD/FACE direction.\\
\ \ - IGNORE the orientation of the torso, shoulders, hips, or feet.\\
3. For VEHICLES, use the usual driving direction / front bumper as the front side.\\
4. For CHAIRS/SOFAS, use the direction a person would face when sitting normally on it.\\
5. For OTHER OBJECTS, use the most functional or visually dominant FRONT side (where logos, screens, or main features appear).\\
6. IGNORE where the object is located in the image (left/right/top/bottom).\\
\ \ Only care about which direction its front/face is pointing relative to the camera.\\
7. You MUST choose exactly ONE label from this list:\\
\ \ [front, front-right, right, back-right, back, back-left, left, front-left].
\end{quote}

\medskip
\noindent\textbf{Prompt Template}
\begin{quote}\ttfamily
Question: In this image, which direction is the \{object\_keyword\}'s FRONT/FACE pointing relative to the camera?\\
Answer with exactly ONE word from the list in \textbackslash boxed\{\}.
\end{quote}

\noindent\textbf{Note:} This strategy prevents potential model hallucination and visual biases from images by strictly defining the meaning of orientation.
\end{tcolorbox}

\subsection{Prompts for Symbolic Geometry Reasoning} \label{Prompts_Geometry_Reasoning}

Following the coordinate transformation $\mathcal{W} \rightarrow \mathcal{F}_{allo}$, all object spatial states are aligned to the target object's \textbf{egocentric frame}. Consequently, during the reasoning phase, we designate the reference object \(O_{ref}\) as ``EGO" to explicitly indicate that the coordinate system has been switched.

\begin{tcolorbox}[vqabox]
\small
\textbf{Prompt: Ego-frame Geometry Context for Symbolic Reasoning.}

\medskip
\noindent\textbf{Geometric Context Definition.}
\begin{quote}\ttfamily
EGO-CENTRIC 3D GEOMETRY CONTEXT\\
EGO-CENTRIC coordinate system definition:\\
- Origin: the ego object's 3D center (ego object = \(O_{ref}\)).\\
- +Z: the ego-facing (forward) direction (estimated/defined by the pipeline).\\
- +X: ego-right; +Y: ego-down.\\[2pt]

For each non-ego object, we provide the ego-frame information\\
coordinates = (x, y, z).\\
Interpretation of signs (ego frame):\\
- x $>$ 0: obj is to the RIGHT of ego; \ \ x $<$ 0: LEFT.\\
- y $>$ 0: obj is DOWN from ego; \ y $<$ 0: UP.\\
- z $>$ 0: obj is in FRONT of ego; \ z $<$ 0: BEHIND.\\

Distance $= \lVert (x,y,z) \rVert_2$ represents the distance from the current object to the ego object. \\[2pt]

Sizes = (dX, dY, dZ) indicate the sizes of the current object along the X, Y, Z axes in the ego coordinate system.


Non-ego objects (ego-frame centers coordinate, distances, and Sizes):\\
Object 1: \{obj\_1\} \textbar\ coordinates $ = (x=\{...\},\,y=\{...\},\,z=\{...\})$
\textbar\ distance=\{...\}  \textbar\ sizes=\{...\}\\
Object 2: \{obj\_2\} \textbar\ coordinates $ = (x=\{...\},\,y=\{...\},\,z=\{...\})$
\textbar\ distance=\{...\} \textbar\ sizes=\{...\}\\
$\cdots$\\[2pt]

\end{quote}

\end{tcolorbox}

\begin{tcolorbox}[vqabox]
\small
\textbf{Prompt: Symbolic Geometry Reasoning over Ego-frame Deltas.}

\medskip
\noindent\textbf{System Instruction.}
\begin{quote}\ttfamily
You are a precise 3D spatial reasoning assistant.\\
You will be given an ego-centric 3D geometry context.\\
Use the geometry context to reason about spatial relationships in ''Question'' \\
\end{quote}

\medskip
\noindent\textbf{Reasoning Prompt.}
\begin{quote}\ttfamily
You FIRST think about the reasoning process as an internal monologue and then provide the final answer. The reasoning process MUST BE enclosed within ⟨think⟩ ⟨/think⟩ tags. The final answer MUST BE put in \textbackslash{boxed}\{\}.
\end{quote}

\medskip
\noindent\textbf{Prompt Template.}
\begin{quote}\ttfamily
\{ego\_geometry\_context\}\\
QUESTION:\\
\{Query \(Q\)\}\\
Reasoning Prompt: \\
\{Reasoning Prompt\

\end{quote}
\end{tcolorbox}

\subsection{Prompts for Symbolic Geometry Reasoning in Camera Frame} \label{Prompts_Geometry_Reasoning_Camera}


Notably, since visual inputs are withheld during the final reasoning phase, the model may struggle to implicitly ground the abstract term "ego" to the observer. Consequently, when designing prompts for the Egocentric Stream, we explicitly employ the terms "Camera" or "Viewer" to designate the origin of the camera coordinate system.

\begin{tcolorbox}[vqabox]
\small
\textbf{Prompt: Camera-frame Geometry Context for Symbolic Reasoning.}

\medskip
\noindent\textbf{Geometric Context Definition.}
\begin{quote}\ttfamily
Camera/Viewer-CENTRIC 3D GEOMETRY CONTEXT\\
Camera/Viewer-CENTRIC coordinate system definition:\\
- Origin: the Camera/Viewer's 3D center.\\
- +Z: the Camera/Viewer-facing (forward) direction.\\
- +X: right; +Y: down.\\[2pt]

For each object, we provide the Camera/Viewer-frame information\\
coordinates = (x, y, z).\\
Interpretation of signs (Camera/Viewer frame):\\
- x $>$ 0: obj is to the RIGHT of Camera/Viewer; \ \ x $<$ 0: LEFT.\\
- y $>$ 0: obj is DOWN from Camera/Viewer; \ y $<$ 0: UP.\\
- z $>$ 0: obj is in FRONT of Camera/Viewer; \ z $<$ 0: BEHIND.\\

Distance $= \lVert (x,y,z) \rVert_2$ represents the distance from the current object to the Camera/Viewer. \\[2pt]

Sizes = (dX, dY, dZ) indicate the sizes of the current object along the X, Y, Z axes in the Camera/Viewer coordinate system.


Objects (Camera/Viewer-frame centers coordinate, distances, and Sizes):\\
Object 1: \{obj\_1\} \textbar\ coordinates $ = (x=\{...\},\,y=\{...\},\,z=\{...\})$
\textbar\ distance=\{...\}  \textbar\ sizes=\{...\}\\
Object 2: \{obj\_2\} \textbar\ coordinates $ = (x=\{...\},\,y=\{...\},\,z=\{...\})$
\textbar\ distance=\{...\} \textbar\ sizes=\{...\}\\
$\cdots$\\[2pt]

\end{quote}

\end{tcolorbox}

\begin{tcolorbox}[vqabox]
\small
\textbf{Prompt: Symbolic Geometry Reasoning over Camera/Viewer-frame Deltas.}

\medskip
\noindent\textbf{System Instruction.}
\begin{quote}\ttfamily
You are a precise 3D spatial reasoning assistant.\\
You will be given a CAMERA/VIEWER-centric 3D geometry context.\\
Use the geometry context to reason about spatial relationships in ''Question'' \\
\end{quote}

\medskip
\noindent\textbf{Reasoning Prompt.}
\begin{quote}\ttfamily
You FIRST think about the reasoning process as an internal monologue and then provide the final answer. The reasoning process MUST BE enclosed within ⟨think⟩ ⟨/think⟩ tags. The final answer MUST BE put in \textbackslash{boxed}\{\}.
\end{quote}

\medskip
\noindent\textbf{Prompt Template.}
\begin{quote}\ttfamily
\{ego\_geometry\_context\}\\
QUESTION:\\
\{Query \(Q\)\}\\
Reasoning Prompt: \\
\{Reasoning Prompt\

\end{quote}
\end{tcolorbox}

\subsection{Prompts for Spatial Query Router} \label{Prompts_Router}

Notably, to facilitate model comprehension, we adopt specific nomenclature in our implementation that maps to the components described in the main text. The pipeline designated as CAMERA\_3D corresponds to the Egocentric Stream, handling reasoning within the camera's coordinate frame. The pipeline labeled EGO\_3D corresponds to the Allocentric Stream, where the system adopts a specific object's egocentric perspective for spatial inference. Additionally, we incorporate an ATTR pipeline designed to focus on object-centric attribute retrieval, such as orientation estimation, existence verification, and standard VQA tasks (e.g., color or material recognition).

\begin{tcolorbox}[vqabox]
\small
\textbf{Prompt R: Spatial Query Router for Pipeline Selection.}

\medskip
\noindent\textbf{System Prompt}
\begin{quote}\ttfamily
You are a routing module for spatial reasoning VQA.\\
Given a single multiple-choice question, choose EXACTLY ONE pipeline label.\\[4pt]
Pipelines:\\
- ATTR:\\
\ \ Use when the question asks for attributes or STATES of objects, including MOST orientation/pose questions.\\
\ \ This INCLUDES:\\
\ \ \ \ (a) single-object facing direction / looking direction / orientation;\\
\ \ \ \ (b) other natural properties of objects(color, material,etc.)\\
- CAMERA\_3D:\\
\ \ Use ONLY when the question requires 3D POSITION reasoning in the CAMERA coordinate system.\\
\ \ When the question asks the position relation of objects from the camera's perspective, it falls under CAMERA\_3D.\\
\ \ If no perspective is specified in the question, the question asks about CAMERA perspective by default.\\
\ \ Examples: which object is closer/farther to the camera, in front of/behind (in 3D), distance comparisons, depth-based relations.\\
- EGO\_3D:\\
\ \ Use ONLY when the question adopts an EGOCENTRIC viewpoint of some object/person and asks\\
\ \ \ \ - The sequence for solving this problem is to first determine the current orientation of A,\\
\ \ \ \ \ \ then assess the position of B relative to A, or thereby identifying which face of A is facing B.\\
\ \ \ \ \ \ This falls under EGO\_3D.\\
\ \ for 3D POSITION relations of other objects relative to that ego coordinate frame.\\
\ \ Examples: ``From X's perspective / If I stand at X facing where it is facing, where is Y relative to me?''\\
Output ONLY one label.
\end{quote}

\medskip
\noindent\textbf{Few-shot Block (ROUTER\_FEWSHOT\_BLOCK).}
\begin{quote}\ttfamily

Example 1:\\
Question: What direction is the girl facing in the image?\\
Label: ATTR\\[4pt]

Example 2:\\
Question: What direction is the girl facing in the image?\\
Label: ATTR\\[4pt]

Example 3:\\
Question: How are the clothes positioned with respect to the cabinet from the camera perspective?\\
Label: CAMERA\_3D

Example 4:\\
Question: If you're looking at the counter, where would you find the table?\\
Label: CAMERA\_3D

Example 5:\\
Question: If I stand at the fridge's position facing where it is facing, is the potted plant to my left or right?\\
Label: EGO\_3D\\[4pt]

Example 6:\\
Question: From the perspective of the boy, where is the dog located?\\
Label: EGO\_3D\\[4pt]

\end{quote}

\medskip
\noindent\textbf{Final Routing Query (assembled in \texttt{build\_router\_prompt}).}
\begin{quote}\ttfamily
Now classify the following question.\\
Question: \{question\}\\
Choose exactly one label from [ATTR, CAMERA\_3D, EGO\_3D].\\

\end{quote}

\end{tcolorbox}


\end{document}